\documentclass{article} 
\usepackage[final]{colm2025_conference} 

\usepackage{microtype}
\usepackage{hyperref}
\usepackage{url}
\usepackage{booktabs}

\usepackage{lineno}

\usepackage[T1]{fontenc}
\usepackage[vietnamese,polish,english,]{babel}
\usepackage{kotex}

\usepackage{graphicx}    
\usepackage{caption}     
\usepackage{subcaption}  
\usepackage{adjustbox} 
\usepackage{multirow}
\usepackage{placeins}
\usepackage{wrapfig}
\usepackage{floatflt} 
\usepackage{amsmath}

\usepackage[listings, skins]{tcolorbox}
\usepackage{listings}
\usepackage{longtable}
\lstdefinelanguage{Markdown}{
  alsoletter={*}, 
  moredelim=[il][\bfseries]{**},
  basicstyle=\ttfamily\footnotesize,
}

\PassOptionsToPackage{table}{xcolor}
\usepackage{colortbl}
\usepackage[table]{xcolor}

\usepackage{pgfmath}
\usepackage{makecell}
\usepackage{fontawesome5}
\usepackage{hyperref}
\usepackage{epigraph}
\usepackage[utf8]{inputenc}
\usepackage{cuted}

\usepackage{fancyvrb}

\usepackage{xspace}

\usepackage[T1]{fontenc}

\definecolor{teal100}{RGB}{0, 90, 90}  
\definecolor{teal90}{RGB}{0, 110, 110} 
\definecolor{teal80}{RGB}{0, 150, 150}  
\definecolor{teal70}{RGB}{0, 220, 220}  

\definecolor{darkblue}{rgb}{0, 0, 0.5}
\hypersetup{colorlinks=true, citecolor=darkblue, linkcolor=darkblue, urlcolor=darkblue}

\definecolor{lightred}{HTML}{e99090}

\newcommand{\name}{\textsc{\small OneRuler}}
\newcommand{\ruler}{\textsc{\small Ruler}}

\newcommand{\github}{\raisebox{-1.5pt}{\includegraphics[height=1.05em]{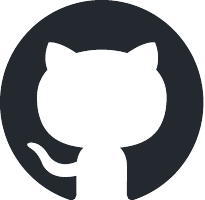}}\xspace}

\title{
  {\includegraphics[height=1.1em]{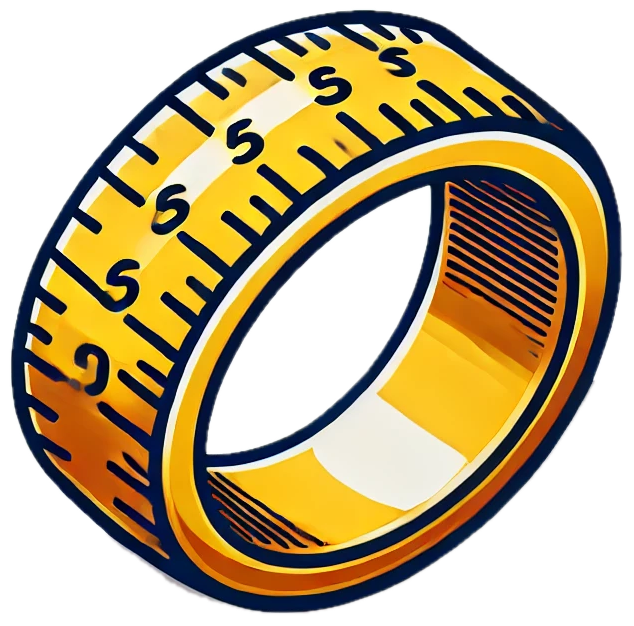}}\
  One ruler to measure them all: \\
  Benchmarking multilingual long-context language models
}

\author{
Yekyung Kim\textsuperscript{\faRuler}, 
Jenna Russell\textsuperscript{\faRuler}, 
Marzena Karpinska\textsuperscript{\faTape},  
Mohit Iyyer\textsuperscript{\faRuler,\faBalanceScale}\\
University of Maryland, College Park\textsuperscript{\faRuler}, 
Microsoft\textsuperscript{\faTape}, 
UMass Amherst\textsuperscript{\faBalanceScale}\\
\texttt{\{yekyung, jennarus, miyyer\}@umd.edu}, 
\texttt{mkarpinska@microsoft.com}
}

\begin{document}

\ifcolmsubmission
\linenumbers
\fi

\maketitle



      


\begin{abstract}
We present \name,
a multilingual benchmark designed to evaluate long-context language models across \textbf{26 languages}. \name\ adapts the English-only \ruler\ benchmark~\citep{hsieh2024ruler} by including seven synthetic tasks that test both retrieval and aggregation, including new variations of the ``needle-in-a-haystack'' task that allow for the possibility of a \emph{nonexistent} needle. We create \name\ through a two-step process, first writing English instructions for each task and then collaborating with native speakers to translate them into 25 additional languages. Experiments with both open-weight and closed language models reveal a widening performance gap between low- and high-resource languages as context length increases from 8K to 128K tokens. Surprisingly, English is not the top-performing language on long-context tasks (ranked 6th out of 26), with Polish emerging as the top language. Our experiments also show that many LLMs (particularly OpenAI's o3-mini-high) incorrectly predict the absence of an answer, even in high-resource languages. Finally, in cross-lingual scenarios where instructions and context appear in different languages, performance can fluctuate by up to 20\% depending on the instruction language. We hope the release of \name\ will facilitate future research into improving multilingual and cross-lingual long-context training pipelines.

\begin{center}
    \renewcommand{\arraystretch}{1.2}
    \begin{tabular}{rl}
         \github & \href{https://github.com/mungg/OneRuler}{\path{https://github.com/mungg/OneRuler}} \\
    \end{tabular}
\end{center}

\end{abstract}

\section{Introduction}
\label{sec:intro}

Long-context language understanding is essential for real-world applications of large language models (LLMs) such as summarization and question answering. However, it is difficult and expensive to conduct realistic evaluations for these tasks \citep{kim2024fables, karpinska-etal-2024-one}, which motivates the use of synthetic benchmarks as proxy diagnostics. One popular example is the ``needle-in-a-haystack'' (NIAH) task \citep{kamradt2023needle}, in which a codeword is inserted into a long document and subsequently queried for. The \ruler\ benchmark \citep{hsieh2024ruler} contains several variants of NIAH (e.g., multiple needles and queries) as well as other synthetic tasks to test aggregation and variable tracing. Unfortunately, \ruler\ and other similar benchmarks mostly test long-context understanding in either just English or in a small number of languages \citep{bai2024longbench, hengle2024multilingualneedlehaystackinvestigating}; as such, it remains unclear how well LLMs perform in \emph{multilingual} and \emph{cross-lingual} long-context scenarios. 

In this paper, we create \name, a multilingual adaptation of \ruler\ that includes seven synthetic tasks (five variants of NIAH as well as two aggregation tasks) in \textbf{26 different languages}, including both low- and high- resource languages. While \ruler\ is intended to test \emph{base} pretrained models, \name\ is intentionally designed for models that have been \emph{post-trained} to follow instructions. Our data collection process involved first writing instructions for all six tasks in English, and then hiring native speakers of the other 25 languages to translate these instructions. Unlike prior work, our NIAH instructions also allow for \textbf{the possibility of a \emph{nonexistent} needle}, where models get credit for identifying that there is no answer. We show that this simple change dramatically lowers the performance of models even on the vanilla NIAH task.

\begin{figure}[t]
    \centering
    \begin{subfigure}[b]{0.49\textwidth}
        \centering
        {\includegraphics[width=\textwidth]{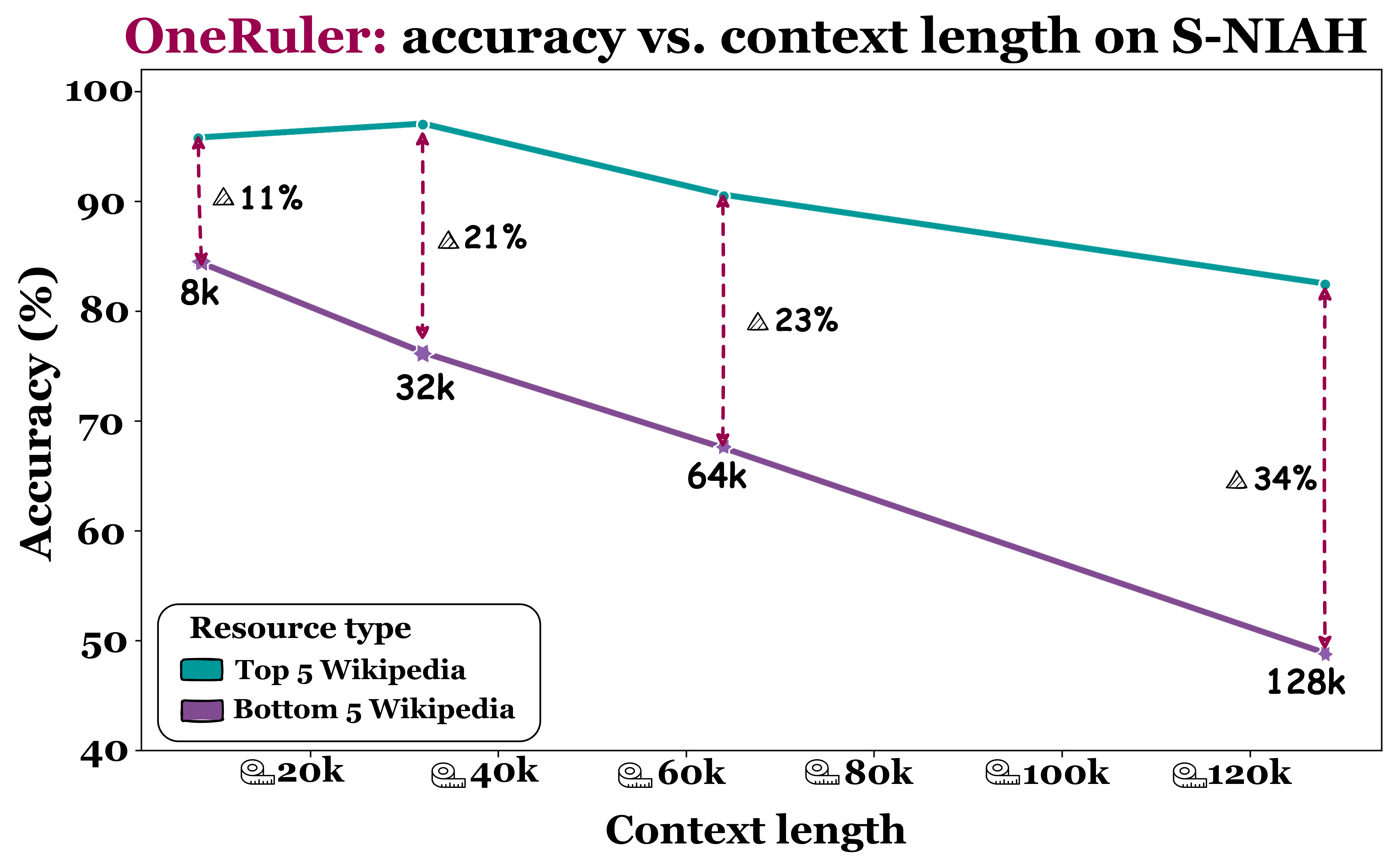}}  
        \caption{S-NIAH Avg. Accuracy}
        \label{fig:sniah_lineplot}
    \end{subfigure}
    \hfill
    \begin{subfigure}[b]{0.49\textwidth}
        \centering
        {\includegraphics[width=\textwidth]{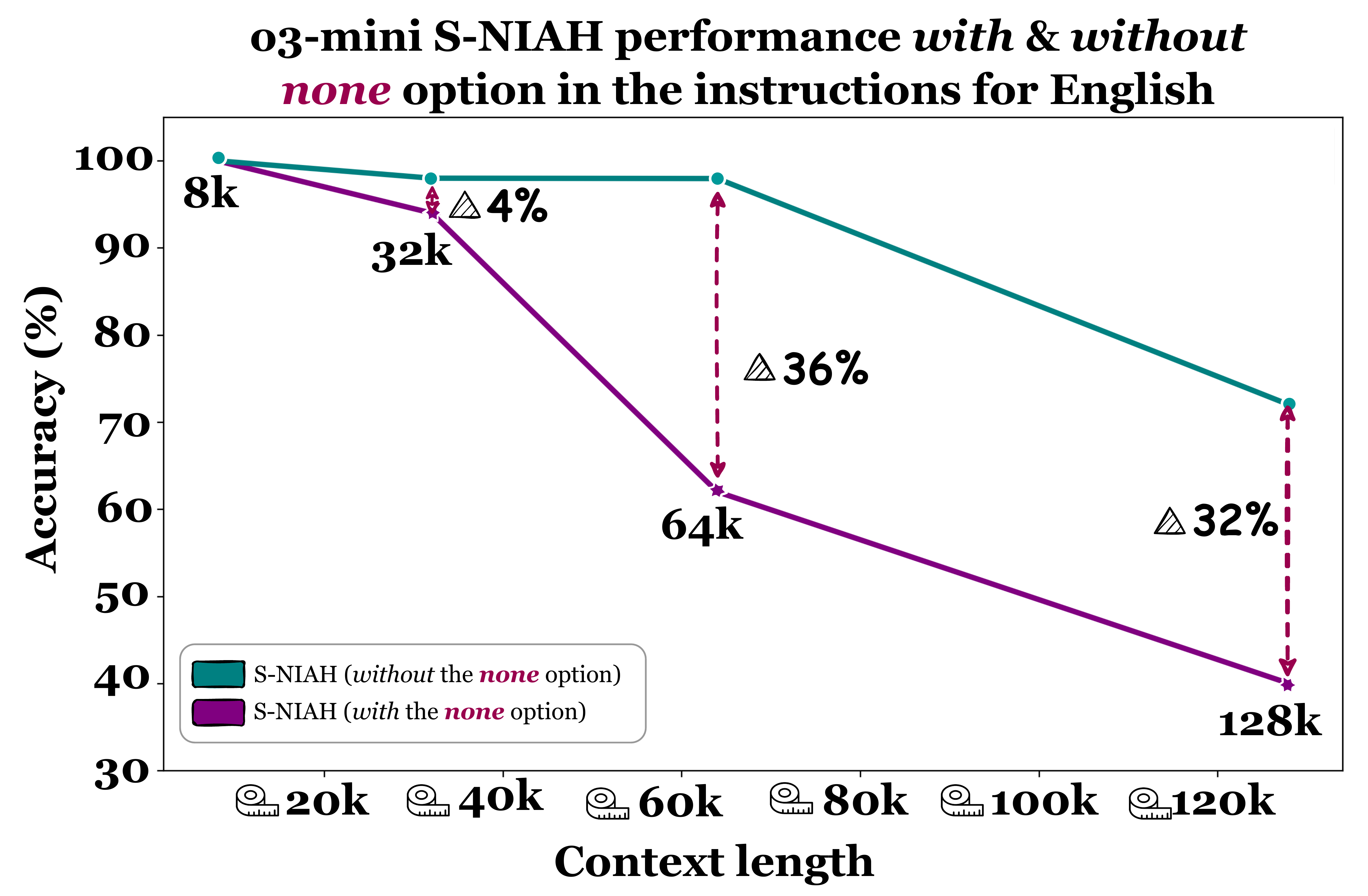}}
    \caption{Impact of nonexistent needle on S-NIAH}
    \label{fig:o3_sniah_none}
    \end{subfigure}
    \caption{(A) Micro-accuracy of all models on the S-NIAH task for the top 5 and bottom 5 languages by Wikipedia size. As context length increases, the performance gap between high-resource and low-resource languages increases. (B) Performance of o3-mini-high on the S-NIAH task in English, with and without the inclusion of the ``None'' option that allows for the possibility of a nonexistent needle. Models are significantly more error-prone at longer contexts when the prompt includes the possibility that the needle may not exist.}
    \label{fig:main}
\end{figure}

We benchmark four recently-released open-weight LLMs of different sizes, Qwen 2.5 (7B and 72B), Llama 3.1 (8B), and Llama 3.3 (70B), as well as two closed-source LLMs (OpenAI's o3-mini-high and Google's Gemini 1.5 Flash). Overall, Gemini 1.5 Flash is the strongest tested model in aggregate, followed by  Qwen 2.5 72B; o3-mini-high, despite its powerful reasoning capabilities, struggles badly on longer contexts.  Interestingly, we observe a widening gap in accuracy (averaged over all tasks and models) between low- and high-resource languages as context length increases (\autoref{fig:main}), suggesting a disparity between languages in long-context pretraining and instruction tuning data. 

Our experiments yield several surprising and counterintuitive results. For one, English is \emph{not} the highest-performing language across all models; in fact, it is the sixth-best
language out of the 26 when evaluated at long-context lengths (64k \& 128k), while Polish takes the top spot.\footnote{Overall, the top-performing language families are Slavic, Romance, and Germanic, while Bantu languages fare poorly despite having over 350M speakers.} 
Also surprising is the fact that even the vanilla NIAH task becomes challenging when the prompt explicitly allows models to respond that the needle is absent, despite near-perfect results observed in \ruler\ and subsequent long-context LLM studies.\footnote{See e.g., Figure 2 of the Qwen 2.5 paper~\citep{qwen_qwen25_2025}, which shows a now-familiar bright green rectangle exhibiting perfect NIAH performance.}
 In fact, a large percentage of errors occur because models incorrectly decide that no needle exists.\footnote{This result is reminiscent of the added challenge posed by SQuAD 2.0's unanswerable questions upon its release~\citep{rajpurkar-etal-2018-know}.} The most difficult task in \name\ is the aggregation task, which requires listing the ten most common words in a long list of words.
Finally, in the \emph{cross-lingual} setting, where the instructions and context are in different languages, we observe that the accuracy can change by up to 20\% depending on the language of instructions.

\section{Creating the \name\ benchmark}
\label{sec:dataset}

\name\ spans seven tasks adapted from \ruler~\citep{hsieh2024ruler}. Five are variants of the needle-in-a-haystack \emph{retrieval} task, differing in the number (and existence) of needles and queries, while the other two require \emph{aggregating} frequent words in a long list. For each task, we evaluate four context lengths (8K, 32K, 64K, 128K) and 26 different languages, with 50 examples per configuration,  totaling 5.2K prompts per task per model.

    \paragraph{Languages:} We include 26 diverse languages: Chinese (zh), Czech (cs), Danish (da), Dutch (nl), English (en), Finnish (fi), French (fr), German (de), Hindi (hi), Hungarian (hu), Italian (it), Japanese (ja), Korean (ko), Norwegian (no), Persian (fa), Polish (pl), Portuguese (pt), Russian (ru), Serbian (sr), Sesotho (st), Spanish (es), Swahili (sw), Swedish (sv), Tamil (ta), Ukrainian (uk), and Vietnamese (vi). These languages provide a solid representation of different language families and writing systems (e.g., Latin, Cyrillic, logographic)
    and exhibit a range of typological features, such as variations in word order and morphological complexity. For fair comparison in retrieval and cross-lingual tasks, we also translated a consistent set of 100 nouns into all 26 languages (see \S\ref{app_sec:data:noun_set} for more details).


\begin{figure}[t]
    \centering
    \resizebox{\linewidth}{!}{    \includegraphics{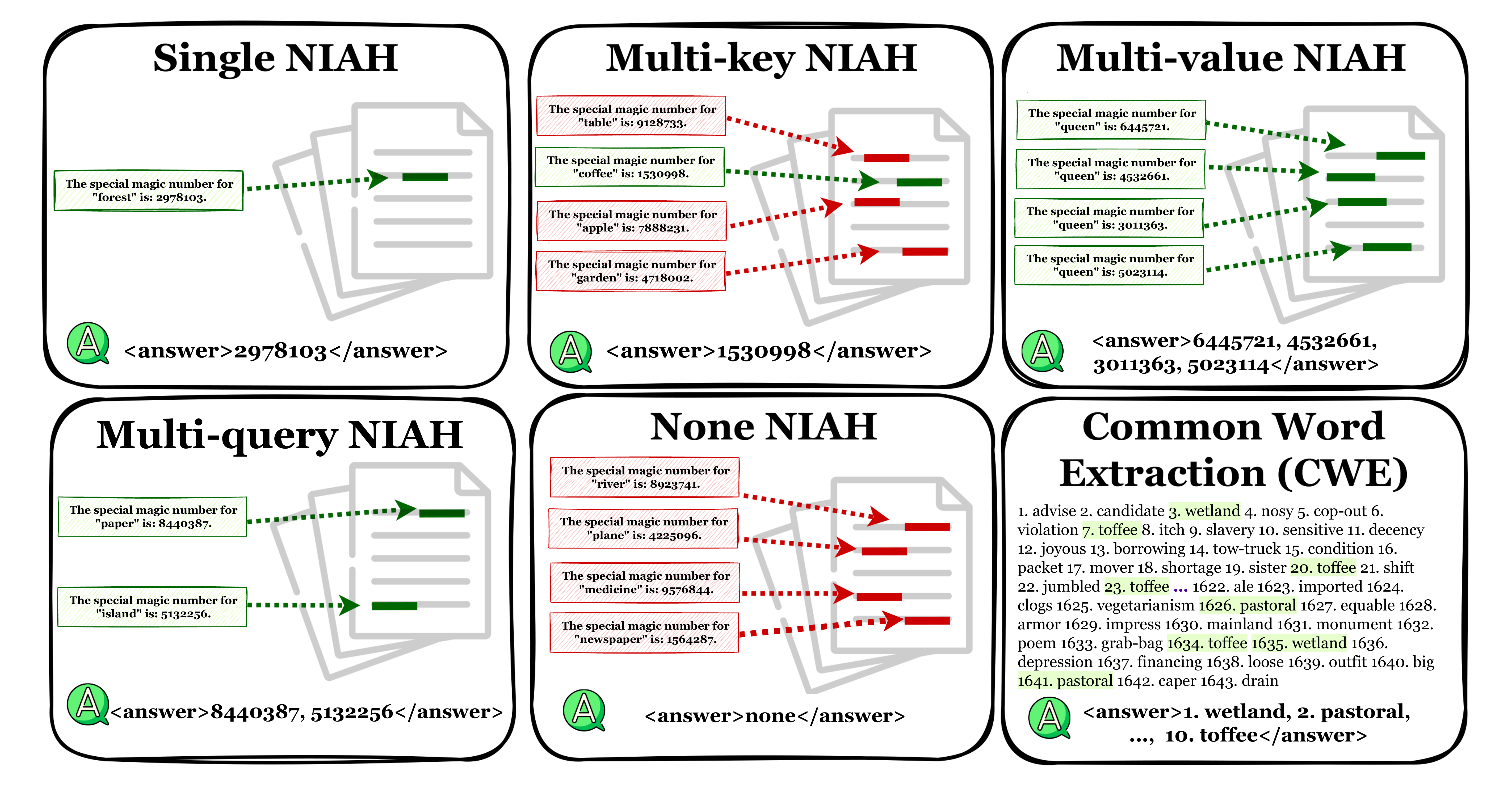}
    }
    \caption{The seven tasks included in \name. Spans highlighted in red are distractors, while green spans contain answers that need to be produced for credit. CWE appears twice (in easy and hard versions with differing frequencies) but shares the same format, hence only one version is shown here. The \textsc{None-NIAH} task is a novel variant in which the needle does not exist in the input context.}
    \label{fig:tasks}
\end{figure}

\paragraph{High vs. low resource languages:} Many of our experiments present comparisons between \emph{high-resource} and \emph{low-resource} languages. To define what constitutes a low-resource language, we rely on the official article count of Wikipedia articles per language \citep{joshi-etal-2020-state, ranathunga-de-silva-2022-languages, Nigatu2024TheZP},\footnote{\url{https://meta.wikimedia.org/wiki/List_of_Wikipedias}} defining a minimum threshold of 250K articles for a language to be considered high resource. Per this definition, we identify four low-resource languages for our study: Hindi, Sesotho, Swahili, and Tamil. 


\paragraph{Translating instructions:} 
As an initial step, we translated English instructions and a list of 100 nouns into 25 languages. For 18 languages, we hired 17 Upwork annotators;\footnote{\url{https://www.upwork.com/}} for the remaining 7 languages, we recruited 6 volunteers from the authors' personal network. 
All annotators were native speakers of the target languages with strong English proficiency.\footnote{Two annotators were native speakers of multiple languages and translated both of those languages (Polish \& Japanese, Russian \& Ukrainian).} 
They were provided with context about the task and its objectives to ensure high-quality translations. Annotators were instructed to translate and localize the instructions to make the prompts sound as natural as possible.\footnote{We pay special attention to the grammar of each language to ensure that any swap of variables will not result in ungrammatical sentences.} They were also instructed to translate 100 nouns based on provided definitions. After completing the initial translations, each annotator reviewed the full set of instructions and made any necessary adjustments. Each annotator was paid \textbf{\$25 USD} per language to translate instructions and 100 nouns, totaling \textbf{\$492 USD}.\footnote{This cost includes contract and processing fees imposed by Upwork. The volunteers were not paid for this task.}


\paragraph{Tokenization:}
It is difficult to conduct a fair comparison across models because they use different tokenizers \citep{ahia-etal-2023-languages}: for example, one of our Tamil documents is 42,124 tokens using Gemini's tokenizer and 103,990 tokens using Qwen's tokenizer. This discrepancy presents us with a choice of either (1) ensuring that the \emph{input text} shown to each model is identical, even if they have differing token counts across models; or (2) ensuring that the \emph{total number of tokens} shown to each model is identical, even if this means some models see more text than others. We decide to report our main results using the second setting to focus specifically on the effect of sequence length on model performance. However, we also report results of experiments run under the first configuration in \S\ref{app_sec:old_experiments}.\footnote{We measure Kendall's $\tau$ over the NIAH tasks across two settings and obtain a coefficient of $0.82$ ($p < 0.001$), indicating strong agreement in model performance rankings.}

\subsection{Retrieval tasks}

We propose five retrieval tasks to assess the model's ability to extract information from extended contexts (see \autoref{fig:tasks}). Each task is based on the needle-in-a-haystack paradigm \citep{kamradt2023needle}, where a target sentence is embedded within a longer text and the model must retrieve specific details. Following \textsc{Ruler} \citep{hsieh2024ruler}, we introduce three variants that modify the number of needles and the amount of information to extract. However, we deviate by reformatting all tasks for \emph{instruction-following} models and also by introducing the possibility of the answer not existing.
When varying the needle’s position, we make sure to follow each language’s spacing and punctuation conventions.
To create plausible contexts for needle injection, we collect and clean 26 open-domain non-copyrighted books, one per language (see \S\ref{app:prompt_templates} for more details). 
Each task is defined as follows:

\begin{itemize}
\item \textbf{Single-NIAH (\textsc{S-Niah}):} This task follows the classic needle-in-a-haystack framework, where a single target sentence (the needle) is embedded in a book-length context. The model must locate this sentence and retrieve the specific number (``value'') associated with the keyword (``key''). In \textsc{S-Niah}, only one needle is present with no distractors. Unlike NIAH configurations in prior work, our prompt template allows for the possibility of a nonexistent needle, even though the needle always exists in \textsc{S-Niah}. This decision (see bolded text in prompt below) reflects real-world scenarios where questions may not always be answerable, and we ablate its impact on performance in \S\ref{sec:analysis}.
\begin{quote}
\scriptsize
\texttt{Please read and memorize the text below. I will ask you about it later.} \\
\texttt{<text>} [CONTEXT] \textit{The special magic number for "[WORD 1]" is: [NUMBER 1].} [CONTEXT] \texttt{</text>} \\
\texttt{<question>} What special magic numbers associated with "[WORD 1]" are mentioned in the provided text? Please list all that apply. \textbf{If no such numbers exist, please answer "none".} \texttt{</question>} \\
Please provide your answer in the following format: \texttt{<answer>}List all numbers here\texttt{</answer>}
\end{quote}


\item \textbf{Multi-key NIAH (\textsc{MK-Niah}):} 
This variant embeds multiple needles with different keys into the context; only one needle contains the correct key. 
Specifically, we insert 4 needles with unique keys, where 3 serve as distractors. The model must identify the needle containing the target key and return its corresponding value.

\item \textbf{Multi-value NIAH (\textsc{MV-Niah}):} In contrast to \textsc{MK-Niah}, this variant inserts 4 needles that share the same key but have different values. 
To successfully complete the task, the model must retrieve all four values associated with the common key.

\item \textbf{Multi-query NIAH (\textsc{MQ-Niah}):} While sharing the same needle structure as \textsc{MK-Niah}, this variant presents multiple queries within each question. The model's response is considered correct only if it accurately retrieves all required information for every query. This tests the model's ability to maintain context awareness across multiple retrieval operations.

\item \textbf{None-NIAH (\textsc{None-Niah}):} This novel variant tests a model's ability to recognize when \emph{no} correct answer exists. The context contains four embedded needles that all function as distractors. This challenges models to acknowledge the absence of a correct response rather than forcing an incorrect selection. The prompt format is identical to \textsc{Single-NIAH}, but the correct answer is always absent.

\end{itemize}

\subsection{Aggregation tasks}

Unlike our retrieval tasks, which focus on extracting specific information from large and irrelevant contexts, aggregation tasks require models to synthesize information across the entire context to generate accurate responses. 
We adapt \ruler's Common Word Extraction (CWE) task, which requires identifying the $n$ most frequent words from a context (see \S\ref{app:word_list_cwe} for more details). Our two CWE settings are: 

\begin{itemize}
\item \textbf{CWE-easy:} The most frequent words in the list appear exactly 30 times each, while other distractor words appear 3 times each. This replicates the parameters from \ruler, chosen because the task proves easy in short context settings but difficult in longer contexts.

\item \textbf{CWE-hard:} We also examine a more difficult setting that changes only the word frequencies. In this setting, the most frequent words appear $20$ times each while distractor words appear $10$ times each. 
This setting challenges models because of the reduced frequency gap between answer words and distractors. 
\end{itemize}


\section{Experiments}
\label{sec:experiments}

\begin{figure*}[t]
    \centering
    \includegraphics[width=1\textwidth]{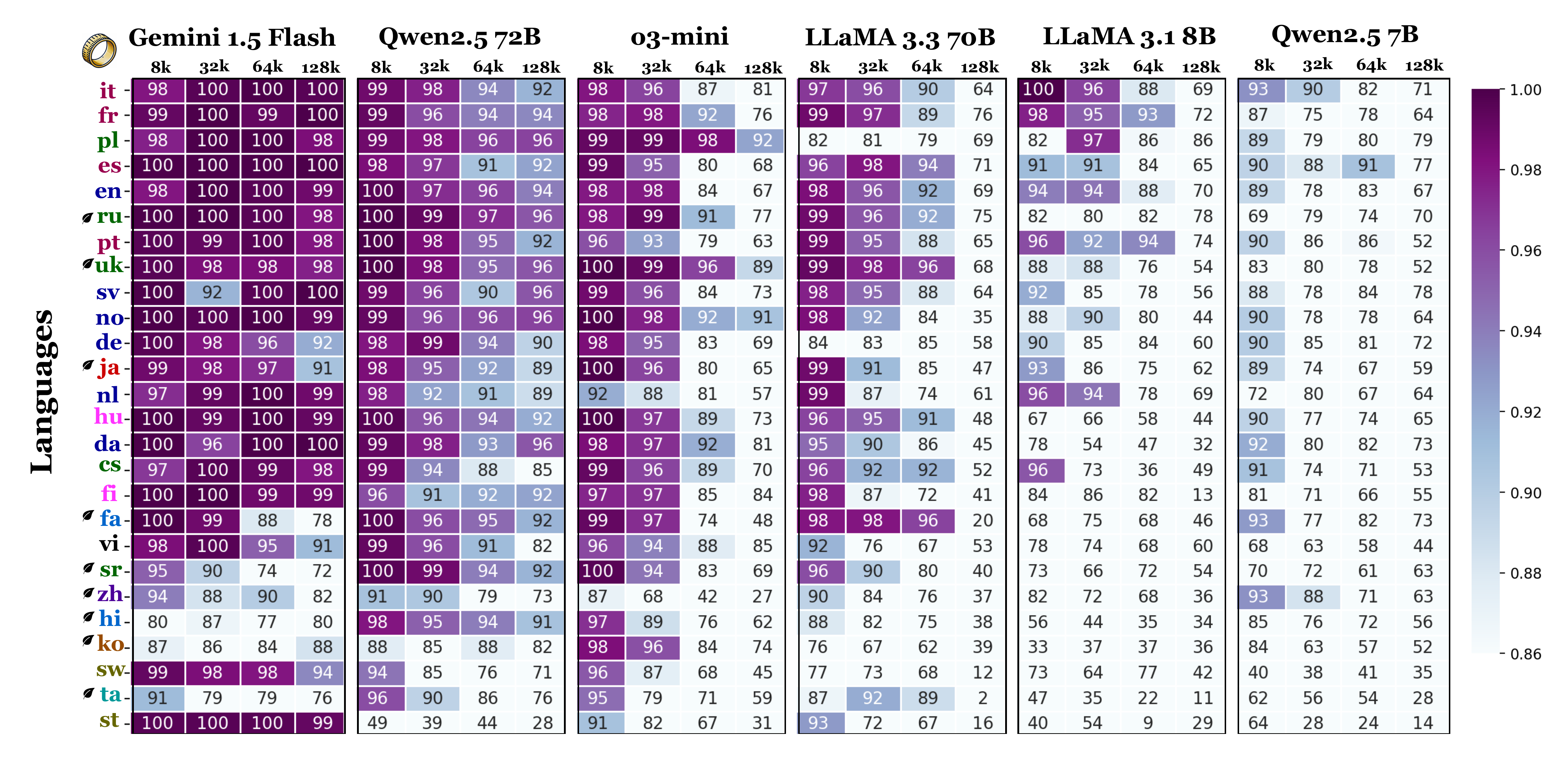}
    \caption{Micro-accuracy across context-lengths and languages for all NIAH tasks. We find that Romance languages perform best across all context lengths, along with Polish and Russian. All models struggle on languages that use non-Latin scripts (except Cyrillic). Gemini-1.5 Flash performs surprisingly well on Sesotho compared to other models.
    }
    \label{fig:heatmap}
\end{figure*}

We evaluate 7 different models on \name\ across four context lengths, reporting accuracy across models, languages, and tasks on the subset of returned responses.{\footnote{For the NIAH task, we discard no-answer cases (2.8\% for o3-mini) and report micro accuracy over the remaining instances. For \textsc{CWE} task, where such cases are more frequent (see \S\ref{app_subsec:additional_reasoning_results}), we treat them as incorrect during evaluation.} While most models perform near perfectly on vanilla NIAH for English at short contexts (8k), accuracies on low-resource languages and those that use non-Latin scripts is drastically lower, especially at longer context lengths. Only Gemini 1.5 Flash and Qwen 2.5 72B perform well on NIAH tasks at long contexts (128K) on aggregate, but they still have room for improvement especially on low resource languages. Our CWE aggregation tasks are difficult for all models, especially the CWE-hard task: none of the test models achieves an accuracy above 1\%.

\paragraph{Model selection:}
We evaluate 5 open-weights models: \citep{deepseekai2025deepseekr1}, Llama 3.3 70B \citep{grattafiori_llama_2024}, Llama 3.1 7B \citep{grattafiori_llama_2024}, Qwen 2.5 \citep{qwen_qwen25_2025} in 7B and 72B variants), and Deepseek-R1,\footnote{Although Deepseek-R1 is an open-weights model, it requires 8 H200-140GB GPUs for inference, which exceeds our available resources. Therefore, we utilized the Fireworks API \url{(https://fireworks.ai/)} for evaluation. Due to cost constraints, we limited our Deepseek-R1 experiments to English.} the latter only for an analysis experiment in English. 
We also compare to two closed-source models: Gemini 1.5 Flash and o3-mini-high. Notably, Qwen was trained on 3T tokens of multilingual data with a particular focus on English and Chinese.
See \S\ref{app_subsec:model_configurations} for more details on model configurations and resources. 

\subsection{Results}

\autoref{fig:lang_performance} shows that \name\ accuracy aggregated over all NIAH tasks and context lengths is (unsurprisingly) higher for high-resource languages than low-resource languages. We do see some correlation between model size and aggregate accuracy on low-resource languages, with the difference in accuracy between high and low resource languages shrinking as model size increases (\autoref{fig:sniah_lineplot}).
We highlight several more interesting findings below:


\begin{figure}[t]
    \centering
    \begin{subfigure}[b]{0.49\textwidth}
        \centering
        {\includegraphics[width=\textwidth]{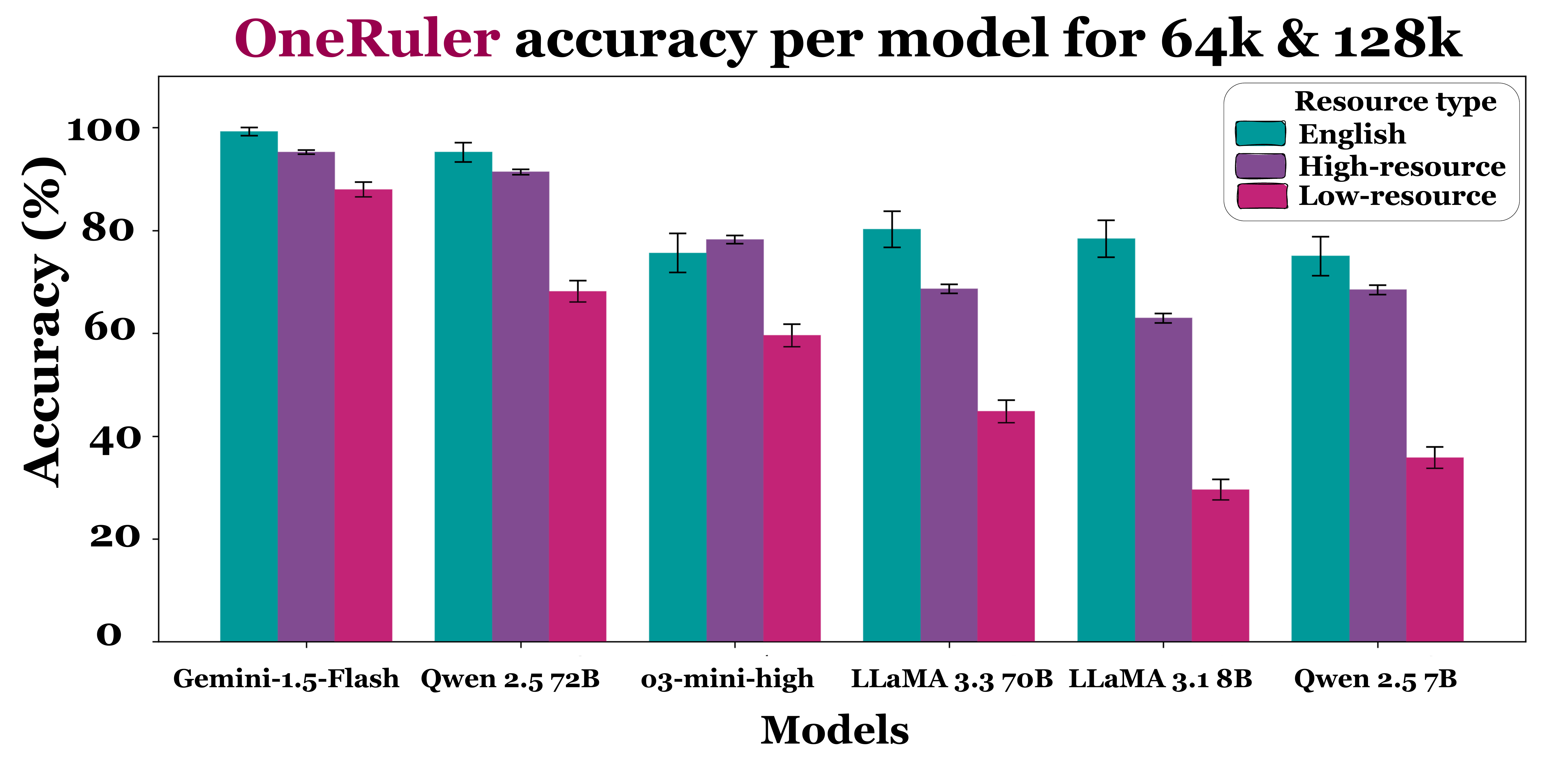}}
        \caption{Performance by model}
        \label{fig:model-resource}
    \end{subfigure}
    \hfill
    \begin{subfigure}[b]{0.49\textwidth}
        \centering
        {\includegraphics[width=\textwidth]{figures/new/bar_langs.pdf }}
    \caption{Performance by language}
    \label{fig:lang_performance}
    \end{subfigure}
    \caption{NIAH performance across models and languages by language resource group for long-context tasks (64K and 128K). Gemini 1.5 Flash demonstrates the best long-context performance, while English and Chinese are surprisingly not among the top five languages.}
    \label{fig:performance_task_model_lang}
\end{figure}

\paragraph{The gap between high- and low-resource languages widens as context size increases:}
As context size increases from 8K to 128K, \autoref{fig:sniah_lineplot} shows that aggregate \name\ accuracy between the top five and bottom five languages by Wikipedia size widens considerably. Specifically, the difference in aggregate accuracy increases from 11\%  with a context length of 8K to 34\% with  context length of 128K. We speculate that the widening gap might be due to a lack of low-resource data used during long context extension \citep{gao2024prolong, lenz2025jamba, grattafiori_llama_2024}: it is possible that long-context capabilities do not easily transfer across languages. 

\paragraph{Low-resource languages are challenging even at short contexts:}
All models demonstrate strong aggregate \name\ accuracy with a context length of 8K, as shown in \autoref{fig:model-resource}. However, they still struggle with low-resource languages like Swahili and Sesotho. This issue is more pronounced in open models, with Llama models exhibiting the most severe performance drops (see \autoref{fig:model_lineplot_niah}). 
This is likely due to LlaMA being predominantly trained on English-centric data \citep{grattafiori_llama_2024}; additionally, the inclusion of the nonexistent needle negatively impacts NIAH task accuracy, as described later in 
\S\ref{sec:analysis}.

\paragraph{English and Chinese are \emph{not} the highest-performing languages:}
English and Chinese dominate the pretraining data of most modern languages, and so we might expect them to be the top-performing languages on \name. However, at context lengths of 64K and 128K, we  unexpectedly observe that Polish is the top performer on NIAH tasks with an average accuracy of 88\% across all models, as depicted in \autoref{fig:lang_performance}. English is only the 6th best language out of the 26, with an average NIAH accuracy of 83.9\%. More shockingly, Chinese is the 4th \emph{worst} language on \name, with an average NIAH accuracy of 62.1\%. While there seems to be some correlation between resource availability and performance (all 4 low-resource languages rank in the bottom 6 languages), it remains unclear why some high-resource languages like Chinese fare worse than anticipated.\footnote{We observe that Qwen's errors on the Chinese S-NIAH task are primarily due to the model frequently generating incorrect 'none' responses. This type of error is not unique to Qwen; it also appears across other models, most notably in o3-mini-high, which exhibits a significant number of such wrong answers (see \S\ref{sec:analysis}).}
In contrast, the top 10 positions are occupied by Slavic, Romance, and Germanic languages, all of which have large Wikipedia size (\autoref{fig:wiki_size}) and use Latin scripts. 

\paragraph{Individual model performance varies:}
\autoref{fig:heatmap} displays the aggregate accuracy of different models on all \name\ NIAH tasks as a function of language and context size. While Gemini 1.5 Flash outperforms all other models across all context lengths, we observe that Qwen 2.5 72B is consistently better than Llama 3.3 70B across all context lengths, with notably higher performance in the 64k and 128k context-length settings. Also interesting is the low average performance of o3-mini-high: it achieves only 67\% accuracy on English at a context length of 128K, compared to 92\% on Polish and 89\% on Ukrainian. 

\begin{wrapfigure}{r}{0.48\columnwidth}
\centering
\includegraphics[width=0.48\columnwidth]{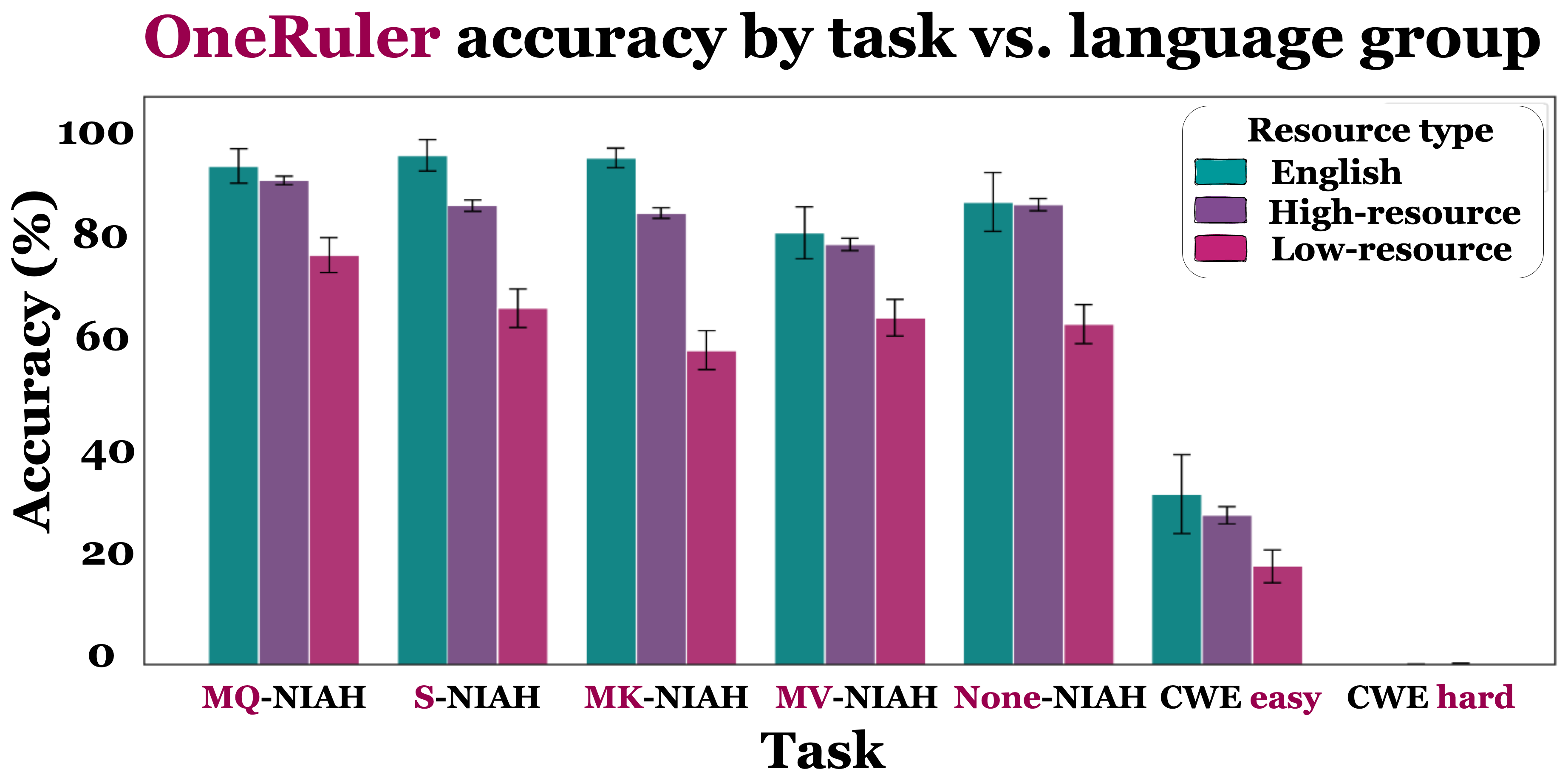}
        \caption{The performance of models on each task, with bars representing English, all other high-resource languages, and low-resource languages.
        }
        \label{fig:task-resource}
        \vspace{-10pt}
\end{wrapfigure}

\paragraph{Models are surprisingly better on multi-query NIAH than single query NIAH for languages other than English: }
\autoref{fig:task-resource} presents task-wise performance. Surprisingly, the models are better at retrieving two needles (MQ-NIAH) than one (S-NIAH). We found that models tend to return 'none' answers more frequently in S-NIAH than in MQ-NIAH, leading to greater performance degradation. We provide further analysis on \textit{nonexistent} needle in \autoref{sec:analysis}.
We also find that MV-NIAH is more challenging than MK-NIAH, possibly because models struggle to retrieve all values associated with a single key or terminate early. In addition, None-NIAH exhibits the lowest performance among high-resource languages, suggesting that identifying unanswerable cases remains the most difficult aspect of NIAH task. 




\paragraph{CWE is much more challenging than NIAH:}


 Compared to the NIAH tasks, on which all models consistently achieve above 80\% average accuracy on high-resource languages,
 the CWE task presents a substantially greater challenge. Average English accuracy over all models is only 31.5\% for the CWE-easy task as shown in \autoref{fig:task-resource}.\footnote{We note that 4 languages (ko, zh, st, sw) have contexts shorter than 128k tokens because the required number of words exceeded our available vocabulary size.}
Three models (Llama 3.3 70B, Qwen 2.5 72B, Gemini 1.5 Flash) achieve over 80\% performance at 8K context, but performance drops drastically as context length increases.
The CWE-hard setting proves unsolvable with nearly 0\% accuracy across all models, indicating that LLMs have significant room for improvement on long-context aggregation tasks.
We further analyze performance across context lengths and models in \S\ref{app_subsec:additional_aggregation_results}.


\section{Analysis}
\label{sec:analysis}

In this section, we dig into some of the surprising results we observe above, seeking to understand what properties of the tasks in \name\ most trouble the models we tested (e.g., nonexistent needles, inefficient reasoning and language-specific issues). We also explore a \emph{cross-lingual} setting in which task instructions and input context are in different languages.

\paragraph{The option to answer \emph{none} makes NIAH significantly harder:}
Since tasks like None-NIAH inherently lack valid answers, we explicitly provided an option for models to respond accordingly by including the instruction: \textit{If no such number exists, please answer `none`}  (\autoref{fig:magic_number_prompt}).
This simple addition made our NIAH tasks much harder than those in \ruler: \autoref{fig:o3_sniah_none} shows that adding this sentence drops S-NIAH accuracy by 32\% 
at a context length of 128K in English. We observe several models, and in particular o3-mini-high as shown in \autoref{fig:o3_none_single}, have a common failure mode of responding \emph{none} when the needle actually exists in the context (see \autoref{app:fig:wrong_none_closed} and 
\autoref{app:fig:wrong_none_opensource} for more detailed analysis). We suspect the inclusion of this sentence may make models overly cautious to responding, and/or many of these models include NIAH data (without the `none` option) during post-training.

\begin{figure}[t]
    \begin{subfigure}[b]{0.42\textwidth}
        \centering
        \includegraphics[width=\textwidth]{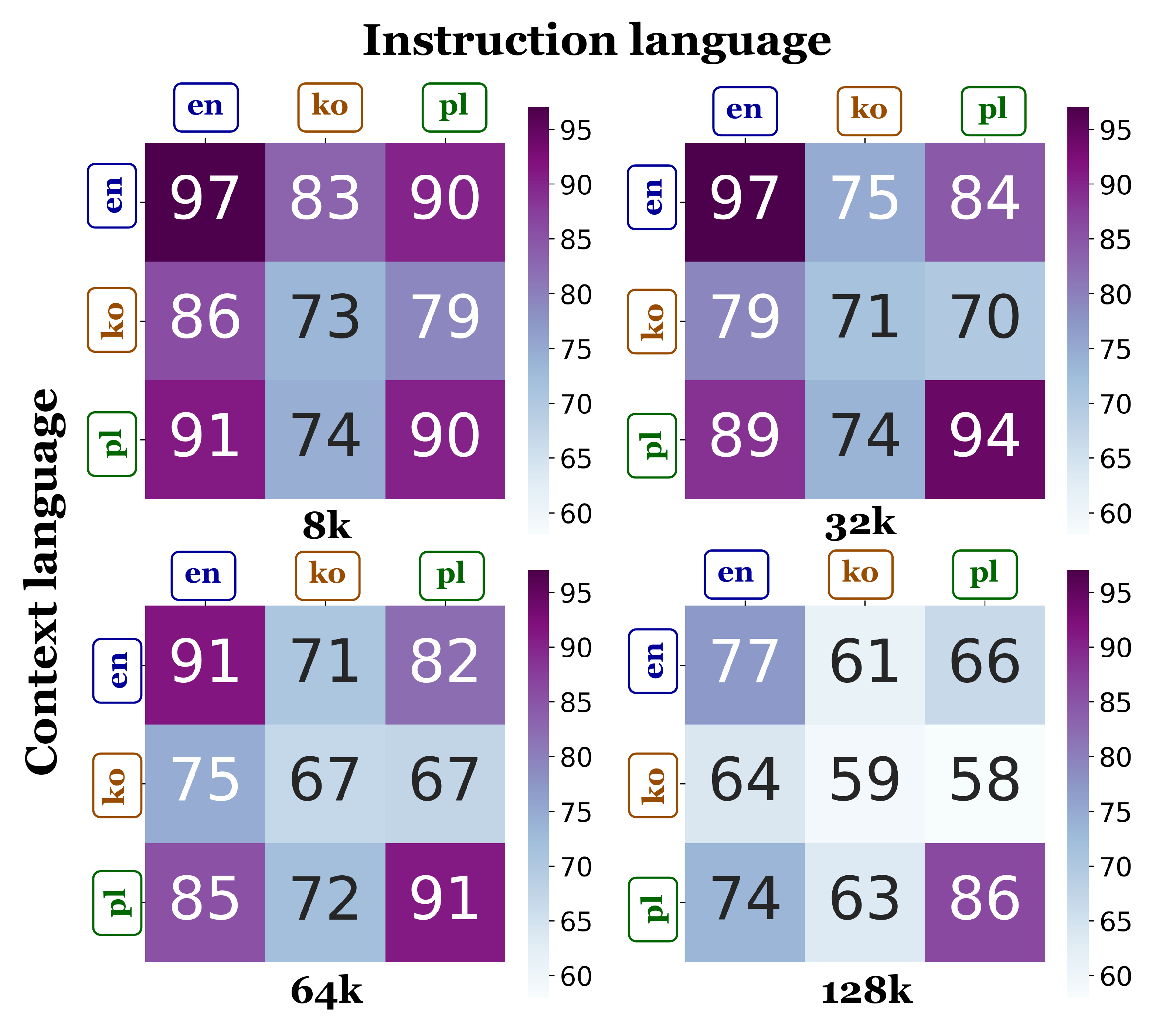}
    \caption{Cross-lingual Performance}
    \label{fig:xling}
    \end{subfigure}
    \hfill
    \begin{subfigure}[b]{0.56\textwidth}
        \centering
        {\includegraphics[width=\textwidth]{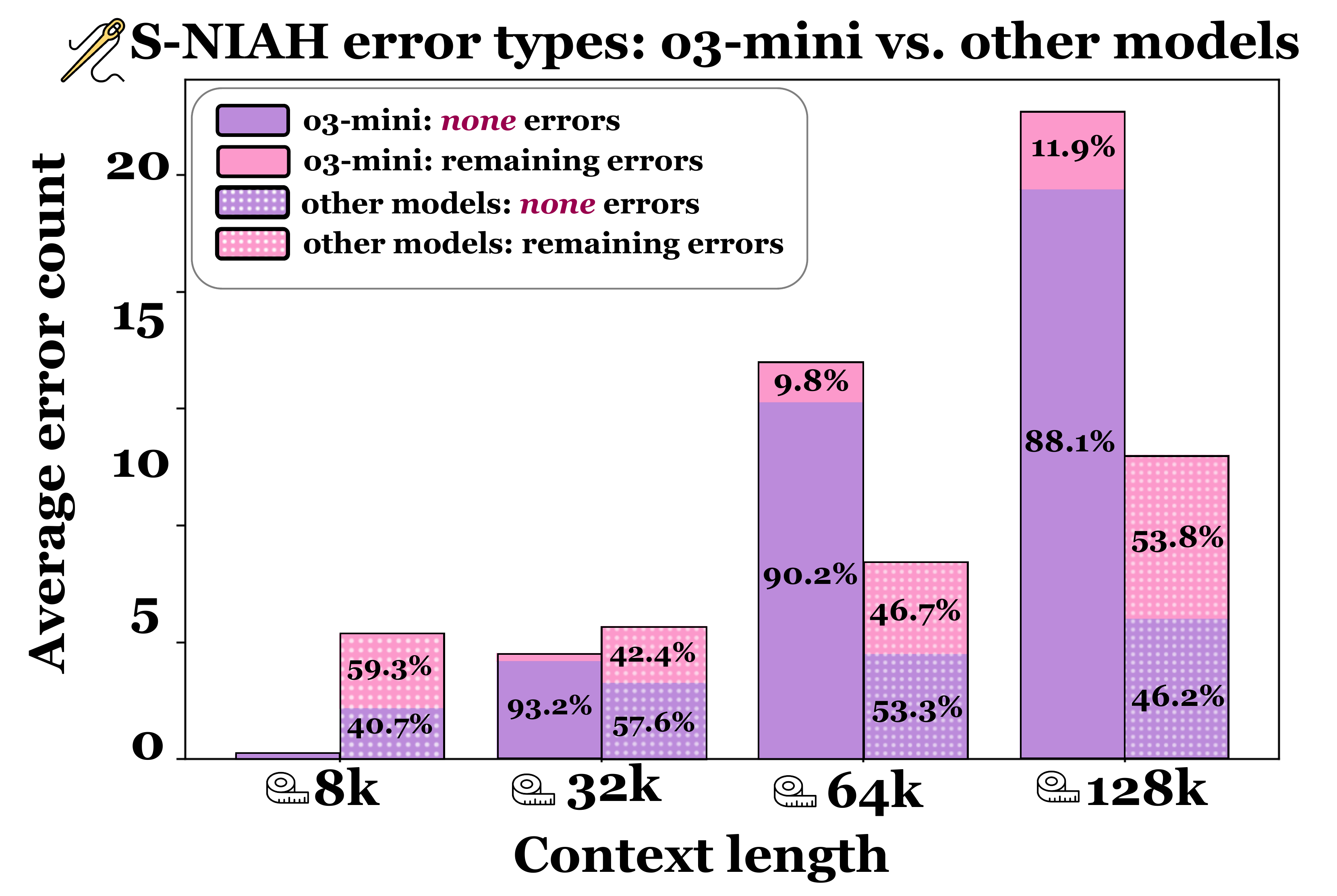}}
    \caption{o3-mini-high Errors in S-NIAH task}
    \label{fig:o3_none_single}
    \end{subfigure}
    \caption{(A) The cross-lingual average accuracy of En, Ko, an Pl on NIAH tasks at each context length. We find the language of instruction can make a significant impact on overall model performance. (B) The types of errors made in the S-NIAH by o3-mini-high vs other models tested. 
    o3-mini-high is more likely to generate an errors than other tasks, and is much more likely to answer `none', despite an answer being present.}
    \label{fig:xling_o3_none}
\end{figure}



\paragraph{Reasoning models behave strangely on NIAH tasks:}
Interestingly, we observe that o3-mini-high produces significantly more reasoning tokens for its incorrect answers than for its correct answers (see \autoref{fig:avg_reasoning_tokens}), suggesting that its reasoning behavior for these simple retrieval tasks is highly inefficient \citep{sui2025stopoverthinkingsurveyefficient}. We also run Deepseek-R1 on the English set of \name\ to see whether it behaves similarly; unfortunately, its high overall performance on English tasks prevents meaningful analysis of reasoning length on incorrect examples. Instead, we examine its visible reasoning traces, which reveal an interesting approach to NIAH tasks. Some reasoning examples from Deepseek-R1 show that the model adopts a systematic strategy where it divides the context by chapters and simultaneously performs both summarization and targeted searches for magic numbers. See §\ref{app_sec:additional_results} for a detailed example.


\paragraph{Cross-lingual performance is language-dependent:}
To test cross-lingual long-context capabilities, we run a variation of \name's NIAH tasks where the instruction language differs from the language of the context and the needles. We run experiments on English, Polish, and Korean (\autoref{fig:xling}), finding that the choice of instruction language significantly impacts performance. Given English contexts, switching the instruction language to a lower resource language such as Korean decreases the average NIAH accuracy across all models at all context lengths; for example at 64K context-length performance drops from 91\% to 71\%. However, if the context is in Korean, switching the instructions to English or Polish \emph{improves} performance: for example, at a context length of 128K, average accuracy increases from 61\% to 77\% when instructions are switched from Korean to English. Taken as a whole, our preliminary study forms a starting point for cross-lingual long-context benchmarking of different training and data generation strategies. 

\paragraph{Complications with CWE:} 
CWE requires models to correctly identify all 10 common words, a task that is trivial for humans but remains surprisingly challenging for LLMs. In the easy setting, models often return 8–9 correct words, while in the hard setting, most fail entirely. High-resource languages tend to perform slightly better, but this advantage diminishes as context length increases (\autoref{fig:cwe_error_analysis}). Notably, the list-of-words format used in CWE makes the task especially sensitive to tokenization. In multilingual settings, tokenizers that produce fewer tokens (e.g., o200k in o3-mini and Gemini) result in a much larger candidate word pool for some languages as shown in \autoref{fig:token_heatmap}, complicating fair comparison across languages. Additionally, reasoning models such as o3-mini-high and Deepseek-R1 often exceed their output token limits (\autoref{fig:reasoning_error_count}). This is largely due to their tendency to recall word lists verbosely. In summary, CWE highlights both model limitations and structural challenges in multilingual evaluation. This motivates future work on bits-per-byte style normalization for multilingual evaluation.

\begin{table*}[t]
\centering
\small
\resizebox{0.99\textwidth}{!}{%
\renewcommand{\arraystretch}{1.6}
\begin{tabular}{p{0.04\textwidth}p{.1\textwidth}p{.13\textwidth}p{0.8\textwidth}p{0.25\textwidth}}
\toprule
\textsc{Lang} & \textsc{Task} & \textsc{Model}  & \textsc{Example} & \textsc{Comment} \\
\midrule

\textbf{DE} & NIAH 
\newline
Single & Qwen 2.5 7B & <Antwort> 1600 </Antwort> Die Zahl 1600 wird im Text erwähnt in Bezug auf die Höhe des Ortes, an dem Joachim Ziemßen lebt: "Sechzehnhundert Meter über dem Meer".
\newline
\textbf{Translation:} The number 1600 is mentioned in the text in relation to the elevation of the place where Joachim Ziemßen lives: "sixteen hundred meters above sea level."
& Used context instead of needle to create answer. \\

\textbf{ZH} & NIAH 
\newline
Single & o3-mini (high) & \texttt{<答案>2</答案> 解析：在文本中有一处写道：“我翻开历史一查……满本都写着两个字是‘吃人’！书上写着这许多字……”这里“书”上所写的内容是“吃人”两个字，因此与“书”相关的魔法数字就是“2”。}
\newline
\textbf{Translation:}  <Answer>2</Answer>
Explanation: In the text, there is a passage that reads, "I flipped through history... the entire book was filled with two characters: 'eat people'! The book contained so many words..." Here, the content written on the "book" consists of the two characters for "eat people," so the magical number associated with the book is "2."
& Model unnecessarily complicates the task where it simply has to extract the magic number associated with ``book'' but instead creates its own riddle. \\

\textbf{PL} & NIAH None & Qwen 2.5 7B & 
<odpowied\'{z}> Brak </od
\texttt{프로그래머:}
\texttt{文字显示不正常，可能是编码问题。请确保使用UTF-8编码打开文件。} 
\newline
\textbf{Translation:} <Answer> None </AnProgrammer: The text is not displaying correctly, possibly due to encoding issues. Please open the file using UTF-8 encoding.
& Model answers then starts producing Korean and switches into Chinese. \\

\textbf{JA} & CWE easy & o3-mini (high) & \texttt{<答え> 在る; 有る, 会議室, けがらわしい, 組合せ, もらう, 出る, 会社, 連休, 年, 物 </答え>}
\newline
\texttt{※なお、こちらの回答は投稿内の全単語リストを機械的に頻度解析した結果（仮想的な解析例）に基づくものであり、実際のデータ処理結果と異なる可能性があります。}
\newline
\textbf{Translation:}
<Answer>
Exist; Have, Conference Room, Filthy, Combination, Receive, Appear, Company, Consecutive Holidays, Year, Object
</Answer>
\newline
※ This answer is based on a mechanical frequency analysis of all words in the post (a hypothetical analysis example) and may differ from actual data processing results.& Model produces a hypothetical answer. \\

\textbf{EN} & CWE easy & Llama 3.1 8B & <Answer>  1. the 
2. and 
3. of 
4. to 
5. a 
6. in 
7. that 
8. is 
9. for 
10. with 
</Answer> & Model returns generic high-frequency words unrelated to the context.
\\
\bottomrule
\end{tabular}%
}
\caption{
Example failures for both NIAH and CWE tasks. Other common cases include returning incomplete answers, partially correct answers, or `none' when the answer exists.}
\label{tab:failures}
\end{table*}

\paragraph{Analysis of common errors:}





For the S-NIAH, models frequently answer `none' (see \autoref{fig:o3_none_single}). Other NIAH tasks are affected by this at lower frequencies, sometimes with numerical responses provided alongside `none' especially if more than one value was requested.
In multi-key and none NIAH, models often return distractors. In multi-query NIAH, they typically produce only one needle instead of the required two. Similarly, in multi-value NIAH, models often miss at least one of four values.
Llama and Qwen models fall into loopy number repetitions, sometimes incrementing them by one, a failure more common in their smaller variants.
In CWE tasks, models frequently return only a subset of the top 10 words, with accuracy declining as context length increases (see \autoref{fig:cwe_error_analysis}).  Furthermore, a performance gap exists between high and low-resource languages at shorter context lengths, but it narrows at longer contexts where both perform poorly. 
Finally, we observe models either hallucinating answers, reformulating the task, or, in the case of Qwen 2.5 7B and LlaMa 3.1 8B, mixing languages almost exclusively for Polish (see \autoref{tab:failures}).






\section{Related work}
\label{sec:related_work}

\paragraph{Evaluation of multilingual long-context LLMs: }
Most related to our work are prior efforts to benchmark multilingual long-context language models.  LongBench \citep{bai2024longbench} includes both synthetic and natural tasks in English and Chinese, while \citet{tanzer2024a} evaluates language models' ability to translate from English to Kalamang, a low-resource language with under 200 speakers. There are also several multilingual variants of NIAH~\citep{hengle2024multilingualneedlehaystackinvestigating,agrawal-etal-2024-evaluating,huang2025benchmax}; however, \name\ includes many more languages than these efforts, in addition to the \emph{none} answer type and evaluation of reasoning models.

\paragraph{Synthetic long-context benchmarks: }
We build on prior synthetic evaluations, most notably \ruler~\citep{hsieh2024ruler}, to benchmark of long-context LLM capabilities. Most of these are largely based on the ``needle-in-a-haystack'' framework \citep{kamradt2023needle}, which has gained popularity due to its ease of evaluation and modification \citep{yuan2024lveval, xu2024stresstesting, song-etal-2025-counting, Laban2024SummaryOA, sharma-etal-2024-losing}. 
Outside of NIAH, the recent LongReason benchmark \citep{ling2025longreason} expands the context of short-context reasoning questions to evaluate long-context capabilities, while GSM-∞\citep{zhou2025gsminfinitellmsbehaveinfinitely} generates long-context tasks with controllable complexity and information density via computational graphs.

\paragraph{Realistic long-context benchmarks:}
While synthetic tasks are cheap and easy to control, they also do not test real-world tasks; as such, other benchmarks (mostly in English) focus on specific tasks such as QA \citep{an2023leval, levy-etal-2024-task}, summarization \citep{kim2024fables} or a suite of many realistic tasks \citep{Shaham2023ZeroSCROLLSAZ, Dong2023BAMBOOAC, li-etal-2024-loogle, Lee2024LongContext, yen2025helmet}. InfiniteBench \citep{zhang-etal-2024-bench} pushed evaluation of context lengths past 100K tokens. Others have proposed evaluation of real-world tasks such as conversations with agents \citep{castillo2024goodai}, and code understanding \citep{repoqa}. BABILong \citep{kuratov2024babilong} and NoCha \citep{karpinska-etal-2024-one} both evaluate reasoning of factuality over long contexts.

\section{Conclusion}
\label{sec:conclusion}

We introduce \name, a synthetic benchmark for multilingual long-context language models across 26 languages that measures both retrieval and aggregation capabilities. Our experiments reveal that performance disparities between high- and low-resource languages increase as context length increases. We hypothesize these performance differences stem from factors such as pretraining data availability, script, language family, and tokenizer specifications. Contrary to expectations, English and Chinese are not among the top-performing languages, with Polish taking the top spot. Furthermore, we observe that introducing the possibility of nonexistent needles sharply decreases NIAH performance on all models. We release \name\ to spur the development of multilingual long-context LLM capabilities.

\section*{Acknowledgments}
We would like to extend our gratitude to the Upwork annotators for their dedicated efforts, and to Ankita Gupta, Chau Pham, Rishanth Rajendhran, Yixiao Song and for voluntarily contributing to the translation. We are also grateful to the members of the UMass NLP and UMD CLIP labs for their valuable feedback. Our deep appreciation goes to Simeng Sun, whose discussions inspired the initial concept of this work and provided many valuable insights. This project was partially supported by awards IIS-2046248, IIS-2312949, and IIS-2202506 from the National Science Foundation
(NSF).

\bibliography{colm2025_conference}
\bibliographystyle{colm2025_conference}

\appendix

\section{Data}
\label{app_sec:data}

In this section we detail the data used for \name. 

\paragraph{Languages}
We use 26 langauges, from a wide range of language families, scripts, and resource sizes. All languages are detailed in \autoref{tab:language_info}. Additionally, we provide a visualization of Wikipedia resource and the number of native speakers in \autoref{fig:wiki_size} and \autoref{fig:speakers_by_family}.

\renewcommand{\arraystretch}{1.2}
\begin{table}[t]
\centering
\resizebox{0.9\textwidth}{!}{%
\footnotesize
\begin{tabular}{l l l l l l l l}
\toprule
\textsc{Language}       & \textsc{ISO} & \textsc{Family}         & \textsc{Subfamily}          & \textsc{Script}              & \textsc{Morph.}        & \textsc{Speakers} & \textsc{Wiki articles} \\
\midrule
English                 & en  & Indo-European  & West Germanic      & Latin               & Analytic      & $\sim$1.5B  & $\sim$6,961,391 \\
German                  & de  & Indo-European  & West Germanic      & Latin               & Fusional      & $\sim$134M  & $\sim$2,992,863 \\
French                  & fr  & Indo-European  & Romance            & Latin               & Fusional      & $\sim$312M  & $\sim$2,668,204 \\
Swedish                 & sv  & Indo-European  & North Germanic     & Latin               & Fusional      & $\sim$10M   & $\sim$2,605,454 \\
Dutch                   & nl  & Indo-European  & West Germanic      & Latin               & Fusional      & $\sim$30M   & $\sim$2,180,999 \\
Russian                 & ru  & Indo-European  & Slavic             & Cyrillic            & Fusional      & $\sim$255M  & $\sim$2,031,560 \\
Spanish                 & es  & Indo-European  & Romance            & Latin               & Fusional      & $\sim$560M  & $\sim$2,013,009 \\
Italian                 & it  & Indo-European  & Romance            & Latin               & Fusional      & $\sim$67M   & $\sim$1,906,293 \\
Polish                  & pl  & Indo-European  & Slavic             & Latin               & Fusional      & $\sim$40M   & $\sim$1,649,832 \\
Chinese                 & zh  & Sino-Tibetan   & Sinitic            & Hanzi               & Analytic      & $\sim$1.1B   & $\sim$1,465,839 \\
Japanese                & ja  & Japonic        & --                 & Kanji/Kana          & Agglutinative & $\sim$125M  & $\sim$1,452,150 \\
Ukrainian               & uk  & Indo-European  & Slavic             & Cyrillic            & Fusional      & $\sim$40M   & $\sim$1,368,238 \\
Vietnamese              & vi  & Austroasiatic  & Vietic             & Latin               & Analytic      & $\sim$86M   & $\sim$1,293,417 \\
Portuguese              & pt  & Indo-European  & Romance            & Latin               & Fusional      & $\sim$264M  & $\sim$1,144,604 \\
Persian (Farsi)         & fa  & Indo-European  & Iranian            & Perso-Arabic        & Fusional      & $\sim$80M   & $\sim$1,030,086 \\
Serbian                 & sr  & Indo-European  & Slavic             & Cyrillic/Latin            & Fusional      & $\sim$12M   & $\sim$703,048  \\
Korean                  & ko  & Koreanic       & --                 & Hangul              & Agglutinative & $\sim$81M   & $\sim$699,221  \\
Norwegian               & no  & Indo-European  & North Germanic     & Latin               & Fusional      & $\sim$5M    & $\sim$643,075  \\
Finnish                 & fi  & Uralic         & Finnic             & Latin               & Agglutinative & $\sim$6M    & $\sim$589,626  \\
Czech                   & cs  & Indo-European  & Slavic             & Latin               & Fusional      & $\sim$10M   & $\sim$563,790  \\
Hungarian               & hu  & Uralic         & Ugric              & Latin               & Agglutinative & $\sim$13M   & $\sim$554,772  \\
Danish                  & da  & Indo-European  & North Germanic     & Latin               & Fusional      & $\sim$6M    & $\sim$306,973  \\
Tamil                   & ta  & Dravidian      & Southern Dravidian & Tamil               & Agglutinative & $\sim$87M   & $\sim$172,122  \\
Hindi                   & hi  & Indo-European  & Indo-Aryan         & Devanagari          & Fusional      & $\sim$600M  & $\sim$165,001  \\
Swahili                 & sw  & Niger--Congo   & Bantu              & Latin               & Agglutinative & $\sim$87M   & $\sim$97,374   \\
Sesotho & st  & Niger--Congo   & Bantu              & Latin               & Agglutinative & $\sim$12M   & $\sim$1,383    \\
\bottomrule
\end{tabular}%
}
\caption{Languages with family, subfamily, script, morphological type, approximate number of speakers (Ethnologue), and number of Wikipedia articles. 
}
\label{tab:language_info}
\end{table}

\begin{table}[htbp]
\centering
\tiny
\begin{tabular}{ll}
\toprule
\textbf{Language} & \textbf{Source} \\
\midrule
Chinese & \url{https://github.com/drkameleon/complete-hsk-vocabulary/tree/main} \\
Czech & \url{https://github.com/gurkylee/Wordlist-Collection/blob/main/languages/czech.txt} \\
Danish & \url{https://github.com/gurkylee/Wordlist-Collection/blob/main/languages/danish.txt} \\
Dutch & \url{https://github.com/gurkylee/Wordlist-Collection/blob/main/languages/dutch.txt} \\
English & Wonderwords library (same as Ruler) \\
Finnish & \url{https://github.com/akx/fi-words/blob/master/words/words.txt} \\
French & \url{https://raw.githubusercontent.com/Blkzer0/Wordlists/refs/heads/master/French.txt} \\
German & \url{https://github.com/Jonny-exe/German-Words-Library} \\
Hindi & \url{https://github.com/eymenefealtun/all-words-in-all-languages/blob/main/Hindi/Hindi.txt} \\
Hungarian & \url{https://github.com/Blkzer0/Wordlists/blob/master/Hungarian.txt} \\
Italian & \url{https://github.com/gurkylee/Wordlist-Collection/blob/main/languages/italian.txt} \\
Japanese & \url{https://github.com/elzup/jlpt-word-list/tree/master} \\
Korean & \url{https://github.com/acidsound/korean_wordlist/blob/master/wordslist.txt} \\
Norwegian & \url{https://github.com/gurkylee/Wordlist-Collection/blob/main/languages/norwegian.txt} \\
Persian (Farsi) & \url{https://github.com/mvalipour/word-list-fa/blob/master/words.txt} \\
Polish & \url{https://github.com/MontrealCorpusTools/sct_resources/blob/main/Polish/words.txt} \\
Portuguese & \url{https://github.com/gurkylee/Wordlist-Collection/blob/main/languages/portuguese.txt} \\
Russian & \url{https://github.com/gurkylee/Wordlist-Collection/blob/main/languages/russian.txt} \\
Serbian & \url{https://github.com/gurkylee/Wordlist-Collection/blob/main/languages/serbian.txt} \\
Southern Sotho & \url{https://github.com/eymenefealtun/all-words-in-all-languages/blob/main/Sesotho/Sesotho.txt} \\
Spanish & \url{https://github.com/gurkylee/Wordlist-Collection/blob/main/languages/spanish.txt} \\
Swahili & \url{https://github.com/michaelnjuguna/All-swahili-words-dictionary/blob/main/kamusi.txt} \\
Swedish & \url{https://raw.githubusercontent.com/martinlindhe/wordlist_swedish/refs/heads/master/swe_wordlist} \\
Tamil & \url{https://github.com/vigneshwaran-chandrasekaran/tamil-language-words-list/blob/master/tamilwords.txt} \\
Ukrainian & \url{https://github.com/gurkylee/Wordlist-Collection/blob/main/languages/ukrainian.txt} \\
Vietnamese & \url{https://github.com/duyet/vietnamese-wordlist} \\
\bottomrule
\end{tabular}
\caption{Sources of wordlist for each language used for the CWE tasks. The lists were downsampled to 10k and part-of-speech tagging was performed using GPT-4o-mini to keep only nouns, verbs, and adverbs.}
\label{tab:language-sources}
\end{table}

\begin{figure}[t]
    \centering
    \resizebox{0.95\columnwidth}{!}{\includegraphics{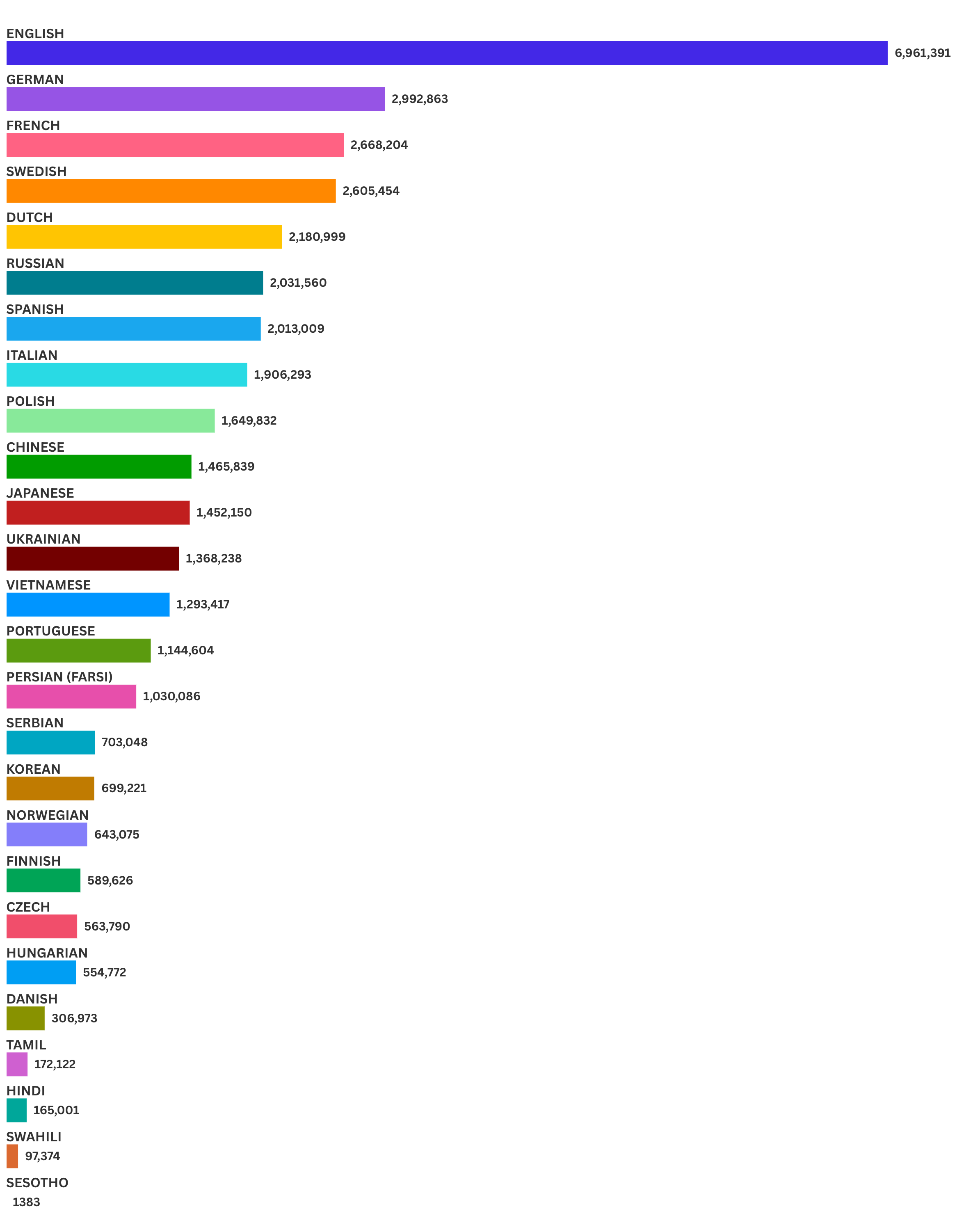}}
    \caption{Language by Wikipedia size. The counts indicate the current number of articles available on Wikipedia for the given language.}
    \label{fig:wiki_size}
\end{figure}

\begin{figure}[t]
    \centering
    \resizebox{0.95\columnwidth}{!}{\includegraphics{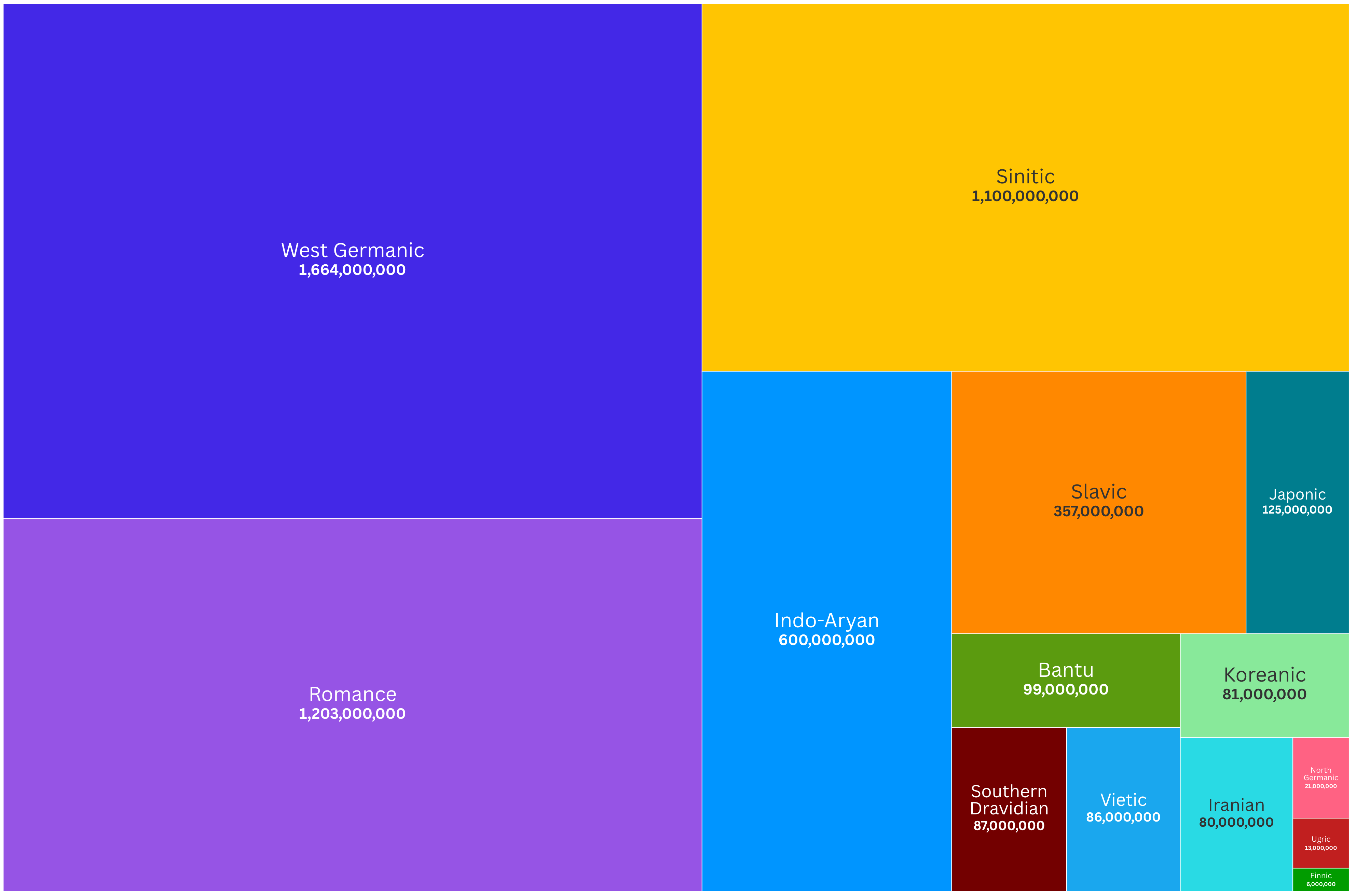}}
    \caption{Language family by the number of speakers as reported by \href{https://www.ethnologue.com/}{Ethnologue}. }
    \label{fig:speakers_by_family}
\end{figure}

\paragraph{Multilingual Noun Set}
\label{app_sec:data:noun_set}
To ensure fair comparison in retrieval and cross-lingual tasks, we selected 100 semantically consistent English nouns and translated them into 26 languages. The initial list of 100 English nouns was generated using the GPT-4o model, along with clear definitions for each word to avoid ambiguity arising from homonyms. Both the English words and their definitions were provided to human annotators, who were instructed to translate the nouns according to the intended meanings.  As a result, we obtained a parallel list of 100 translated nouns across 26 languages.
Each language was primarily handled by a single native speaker responsible for translating all prompts and noun lists in that language. In a few cases, a second annotator reviewed the translations for quality assurance, but each data point was authored by a single individual.

\paragraph{Book Data}
For NIAH tasks, we fill the surrounding context with books from each respective language. Each book was processed to remove the front and back matter. In \autoref{tab:book_info}, we detail for each book the title, author, year, and original language of publication. All books must be long enough to fill the 128k context-length tasks for the specified language. For shorter books, we replicate part of the book to fit the context. Most of the books are older due to the copyright restrictions, and we acknowledge that their age and linguistic style might have influenced models' performance. However, this relationship is not straightforward (i.e., the performance on newer books is not necessarily better than on older books).


\begin{table*}[t]
    \centering
    \resizebox{\linewidth}{!}{ 
    \begin{tabular}{clllc}
        \hline
        \textbf{Lang.} & \textbf{Translated Title (Original Title)} & \textbf{Author} & \textbf{Year} & \textbf{Token Count} \\
        \hline
        cs & \textit{The Good Soldier Švejk, Vol.3 (Osudy dobrého vojáka Švejka za světové války)} & Jaroslav Hašek & 1923 & 174576 \\
        da & \textit{Elderflower and Hops (Hyld og Humle: Fortællinger)} & Sophie Breum & 1900 & 76082 \\
        de & \textit{The Magic Mountain (Der Zauberberg)} & Thomas Mann & 1924 & 598161 \\
        en & \textit{Little Women} & Louisa May Alcott & 1868 & 233668 \\
        es & \textit{Don Quixote} & Miguel de Cervantes & 1605 & 626105 \\
        fa & \textit{The Blind Owl}  & Sadegh Hedayat & 1936 & 77451 \\ 
        fi & \textit{Sylvi} & Minna Canth & 1893 & 60803 \\
        fr & \textit{Les Misérables} & Victor Hugo & 1862 & 970526 \\
        hi & \textit{Chandrakanta} & Devaki Nandan Khatri & 1888 & 813743 \\
        hu & \textit{King Midas, Vol. 1 (Midás király)} & Zoltán Ambrus & 1891 & 224484 \\
        it & \textit{The Late Mattia Pascal} & Luigi Pirandello & 1904 & 144402 \\
        ja & \textit{Kokoro} (\texttt{こころ}) & Natsume Sōseki & 1914 & 185905 \\
        ko & \textit{A Boy's Sorrow (소년의 비애)} & Yi Kwang-su & 1917 & 12279 \\
        nl & \textit{The Diary of a Young Girl (Het Achterhuis)} & Anne Frank & 1947 & 136314 \\
        no & \textit{Kristin Lavransdatter, Vol. 2: The Wife} & Sigrid Undset & 1921 & 251016 \\
        pl & \textit{Nights and Days, Vol. 3 (Noce i dnie)} & Maria Dąbrowska & 1934 & 452847 \\
        pt & \textit{The Book of Disquiet (Livro do Desassossego)} & Fernando Pessoa & 1982 & 248245 \\
        ru & \textit{War and Peace (Война и миръ)} & Leo Tolstoy & 1869 & 1112514 \\
        sr & \textit{Seconds of Eternity (Sekund večnosti, istočnjački roman)} & Dragutin Ilić & 1921 & 45038 \\
        st & \textit{Chaka} & Thomas Mofolo & 1925 & 119342 \\
        sv & \textit{The Story of Gösta Berling} & Selma Lagerlöf & 1891 & 255416 \\
        sw & \textit{My Life Fifty Years After (Maisha yangu na baada y Miaka Hamsini)} & Shaaban Robert & 1958 & 61490 \\
        ta & \textit{Ponniyin Selvan, Vol. 1: The First Floods} & Kalki Krishnamurthy & 1950 & 912523 \\
        uk & \textit{After Finishing School (Instytutka)} & Marko Vovchok & 1862 & 41056 \\
        vi & \textit{Pure Heart (Tố tâm)} & Ho\`ang Ng\d{o}c Ph\'ach & 1925 & 72267 \\
        zh & \textit{Call to Arms (吶喊)} & Lu Xun & 1922 & 153415 \\
        \hline
    \end{tabular}
    }
        \caption{Complete list of books used for needle injections in the retrieval task. Each row contains the language the book was originally published in, the title, author name, published year, and token count. Tokens counts calculated using tiktoken (o200k). }
    \label{tab:book_info}
\end{table*}

\paragraph{Word lists used for the CWE task in 26 languages}
\label{app:word_list_cwe}
We extracted word lists for 26 languages from GitHub repositories (\autoref{tab:language-sources}) dedicated to language wordlists, using part-of-speech tagging to identify nouns, adjectives, and verbs. 

\section{Generations}

\paragraph{Prompt Templates}
\label{app:prompt_templates}
FFor each task, we use consistent templates with minimal rewording. Translations were done at the instruction level, with each paragraph translated separately. Translators were informed of the task's purpose and the words to be replaced, ensuring the wording allowed for `key` substitution without affecting grammar The correct instructions and words were added to make the final prompts. The English versions of the prompts are as follows: (1) the Single-Niah task (\autoref{fig:magic_number_prompt}), (2) the multi-query task (\autoref{fig:magic_number_prompt_multiquery}), (3) the multi-key task (\autoref{fig:magic_number_prompt_multikey}), (4) the none-NIAH task (\autoref{fig:magic_number_prompt_none}), (5) the multi-value task (\autoref{fig:magic_number_prompt_multivalue}), and (6) the single NIAH task without the inclusion of none \autoref{fig:magic_number_prompt_no_none} (used for ablation study). The CWE prompt is shown in (\autoref{fig:cwe_prompt}). We include the English prompts for understanding, but for experiments translated instruction prompts are used in the language of the context (except the cross-lingual setup). We ensured all variables could be substituted in templates without grammatical changes (e.g., singular/plural forms, gender variants, or grammatical cases). All instructions and needles are verified translations by native speakers of each respective language. 

\begin{figure*}[t]
\centering
\begin{tcolorbox}[colback=violet!10!white, colframe=violet, title=\centering\footnotesize Single NIAH Template]
\lstset{
    basicstyle=\ttfamily\footnotesize,
    breaklines=true,
    frame=none,
    xleftmargin=0pt,
    framexleftmargin=0pt,
    columns=fullflexible,
    tabsize=1,
    breakindent=0pt,
    breakautoindent=false,
    postbreak=\space,
    showstringspaces=false,
}
\begin{lstlisting}[
    language=Markdown
]
Please read and memorize the text below. I will ask you about it later.
<text>
[CONTEXT]
The special magic number for "[WORD 1]" is: [NUMBER 1].
[CONTEXT]
</text>

<question>
What special magic numbers associated with "[WORD 1]"  are mentioned in the provided text? Please list all that apply. **If no such numbers exist, please answer "none".
</question>

Please provide your answer in the following format:
<answer>List all numbers here</answer>

\end{lstlisting}
\end{tcolorbox}
\caption{Prompt template for our ``Single NIAH'' task with one magic number. The bolded sentence introduces the possibility of no answer existing, which we show significantly hurts model accuracies, even for this Single NIAH task where the answer always exists.}
\label{fig:magic_number_prompt}
\end{figure*}

\begin{figure*}[t]
\centering
\begin{tcolorbox}[colback=violet!10!white, colframe=violet, title=Multi-Key NIAH Template]
\lstset{
    basicstyle=\ttfamily\footnotesize,
    breaklines=true,
    frame=none,
    xleftmargin=0pt,
    framexleftmargin=0pt,
    columns=fullflexible,
    tabsize=1,
    breakindent=0pt,
    breakautoindent=false,
    postbreak=\space,
    showstringspaces=false,
}
\begin{lstlisting}[language=Markdown]
Please read and memorize the text below. I will ask you about it later
<text>
[CONTEXT]
The special magic number for "[WORD 1]" is: [NUMBER 1].
[CONTEXT]
The special magic number for "[WORD 2]" is: [NUMBER 2].
[CONTEXT]
The special magic number for "[WORD 3]" is: [NUMBER 3].
[CONTEXT]
The special magic number for "[WORD 4]" is: [NUMBER 4].
</text>

<question>What special magic numbers associated with "[WORD 3]"  are mentioned in the provided text? Please list all that apply. If no such numbers exist, please answer "none".</question>

Please provide your answer in the following format:
<answer>List all numbers here</answer>

\end{lstlisting}
\end{tcolorbox}
\caption{Prompt template for Multi-Key NIAH Task with multiple magic numbers.}
\label{fig:magic_number_prompt_multikey}
\end{figure*}

\begin{figure*}[htbp]
\centering
\begin{tcolorbox}[colback=violet!10!white, colframe=violet, title=Multi-Value NIAH Template]
\lstset{
    basicstyle=\ttfamily\footnotesize,
    breaklines=true,
    frame=none,
    xleftmargin=0pt,
    framexleftmargin=0pt,
    columns=fullflexible,
    tabsize=1,
    breakindent=0pt,
    breakautoindent=false,
    postbreak=\space,
    showstringspaces=false,
}
\begin{lstlisting}[language=Markdown]
Please read and memorize the text below. I will ask you about it later
<text>
[CONTEXT]
The special magic number for "[WORD 1]" is: [NUMBER 1].
[CONTEXT]
The special magic number for "[WORD 1]" is: [NUMBER 2].
[CONTEXT]
The special magic number for "[WORD 1]" is: [NUMBER 3].
[CONTEXT]
The special magic number for "[WORD 1]" is: [NUMBER 4].
</text>

<question>What special magic numbers associated with "[WORD 1]" are mentioned in the provided text? Please list all that apply. If no such numbers exist, please answer "none".</question>

Please provide your answer in the following format:
<answer>List all numbers here</answer>

\end{lstlisting}
\end{tcolorbox}
\caption{Prompt template for Multi-Value NIAH Task with multiple magic numbers. }
\label{fig:magic_number_prompt_multivalue}
\end{figure*}

\begin{figure*}[htbp]
\centering
\begin{tcolorbox}[colback=violet!10!white, colframe=violet, title=Multi-Query NIAH Template]
\lstset{
    basicstyle=\ttfamily\footnotesize,
    breaklines=true,
    frame=none,
    xleftmargin=0pt,
    framexleftmargin=0pt,
    columns=fullflexible,
    tabsize=1,
    breakindent=0pt,
    breakautoindent=false,
    postbreak=\space,
    showstringspaces=false,
}
\begin{lstlisting}[language=Markdown]
Please read and memorize the text below. I will ask you about it later
<text>
[CONTEXT]
The special magic number for "[WORD 1]" is: [NUMBER 1].
[CONTEXT]
The special magic number for "[WORD 2]" is: [NUMBER 2].
</text>

<question>What special magic numbers associated with "[WORD 1]" and "[WORD 2]"  are mentioned in the provided text? Please list all that apply. If no such numbers exist, please answer "none".</question>

Please provide your answer in the following format:
<answer>List all numbers here</answer>

\end{lstlisting}
\end{tcolorbox}
\caption{Prompt template for Multi-Query NIAH Task with multiple magic numbers.}
\label{fig:magic_number_prompt_multiquery}
\end{figure*}

\begin{figure*}[htbp]
\centering
\begin{tcolorbox}[colback=violet!10!white, colframe=violet, title=None NIAH Template]
\lstset{
    basicstyle=\ttfamily\footnotesize,
    breaklines=true,
    frame=none,
    xleftmargin=0pt,
    framexleftmargin=0pt,
    columns=fullflexible,
    tabsize=1,
    breakindent=0pt,
    breakautoindent=false,
    postbreak=\space,
    showstringspaces=false,
}
\begin{lstlisting}[language=Markdown]
Please read and memorize the text below. I will ask you about it later
<text>
[CONTEXT]
The special magic number for "[WORD 1]" is: [NUMBER 1].
[CONTEXT]
The special magic number for "[WORD 2]" is: [NUMBER 2].
[CONTEXT]
The special magic number for "[WORD 3]" is: [NUMBER 3].
[CONTEXT]
The special magic number for "[WORD 4]" is: [NUMBER 4].
</text>

<question>What special magic numbers associated with "[WORD 5]"  are mentioned in the provided text? Please list all that apply. If no such numbers exist, please answer "none".</question>

Please provide your answer in the following format:
<answer>List all numbers here</answer>

\end{lstlisting}
\end{tcolorbox}
\caption{Prompt template for None NIAH Task with no matching magic numbers.}
\label{fig:magic_number_prompt_none}
\end{figure*}

\begin{figure*}[htpb]
\centering
\begin{tcolorbox}[colback=violet!10!white, colframe=violet, title=\centering Single NIAH - No 'None' Template]
\lstset{
    basicstyle=\ttfamily\footnotesize,
    breaklines=true,
    frame=none,
    xleftmargin=0pt,
    framexleftmargin=0pt,
    columns=fullflexible,
    tabsize=1,
    breakindent=0pt,
    breakautoindent=false,
    postbreak=\space,
    showstringspaces=false,
}
\begin{lstlisting}[language=Markdown]
Please read and memorize the text below. I will ask you about it later.
<text>
[CONTEXT]
The special magic number for "[WORD 1]" is: [NUMBER 1].
[CONTEXT]
</text>

<question>
What special magic numbers associated with "[WORD 1]"  are mentioned in the provided text? Please list all that apply. **If no such numbers exist, please answer "none".
</question>

Please provide your answer in the following format:
<answer>List all numbers here</answer>

\end{lstlisting}
\end{tcolorbox}
\caption{Prompt template for Single NIAH Task, without the inclusion of None in the prompt. }
\label{fig:magic_number_prompt_no_none}
\end{figure*}


\begin{figure*}[htpb]
\centering
\begin{tcolorbox}[colback=violet!10!white, colframe=violet, title=CWE Template]
\lstset{
    basicstyle=\ttfamily\footnotesize,
    breaklines=true,
    frame=none,
    xleftmargin=0pt,
    framexleftmargin=0pt,
    columns=fullflexible,
    tabsize=1,
    breakindent=0pt,
    breakautoindent=false,
    postbreak=\space,
    showstringspaces=false,
}
\begin{lstlisting}[language=Markdown]
Below is a numbered list of words. In these words, some appear more often than others. Memorize the ones that appear most often.
<list>[ENUMERATED LIST OF WORDS]</list>
<question> What are the 10 most common words in the list above? </question>
Please provide your answer in the following format
<answer> List the words here </answer>
\end{lstlisting}
\end{tcolorbox}
\caption{Prompt template for Most Common Words Task.}
\label{fig:cwe_prompt}
\end{figure*}

\begin{table*}[t!]
    \centering
    \scriptsize		
    \begin{tabular}{lccccc}
        \toprule
        \textbf{Model}  & \textbf{Size} & \textbf{Context Length} & \textbf{Huggingface \cite{DBLP:journals/corr/abs-1910-03771} / API} & \textbf{Cost} \\
        \midrule
        Gemini 1.5 Flash \citep{team_gemini_2024}  & - & 1M & gemini-1.5-flash & \$137.7 \\
        o3-mini-high \citep{openai2025o3mini} & - & 200K & o3-mini-2025-01-31 & \$1932.5 \\
        \midrule
        Deepseek-R1 \citep{deepseekai2025deepseekr1} & 685B & 128k & deepseek-r1 & \$333.2\\
        Llama3.3 \citep{grattafiori_llama_2024}  & 70B & 128K & meta-llama/Llama-3.3-70B-Instruct & - \\
        Llama3.1 \citep{grattafiori_llama_2024}  & 8B & 128K & meta-llama/Meta-Llama-3.1-8B-Instruct & - \\
        Qwen2.5-72B \citep{qwen_qwen25_2025}  & 72B & 128K & Qwen/Qwen2.5-72B-Instruct & - \\
        Qwen2.5-7B \citep{qwen_qwen25_2025}  & 7B & 128K & Qwen/Qwen2.5-7B-Instruct &  \\
        \bottomrule
    \end{tabular}
    \caption{Comparison of Open and Closed Language Models used in experiments}
    \label{tab:model_comparison_appendix}
\end{table*}

\subsection{Model Configurations}
\label{app_subsec:model_configurations}
For o3-mini, we set reasoning efforts to \textit{high} and configured the maximum output token length to 10k, with temperature fixed at 1.0 as per OpenAI defaults. 
For Deepseek-R1, we set the maximum output token length to 8k and used a temperature of 0.6.
\begin{wrapfigure}{r}{0.5\textwidth}
\vspace{-10pt}
    \centering
    \scriptsize
    \setlength{\tabcolsep}{4pt}
    \resizebox{0.5\textwidth}{!}{ 
    \begin{tabular}{lccc}
        \toprule
        \textbf{Model}  & \textbf{\# Param} & \textbf{Ctx len} & \textbf{\# Vocab} \\
        \midrule
        \faLock\ \href{https://ai.google.dev/gemini-api/docs/models/gemini\#gemini-1.5-flash}{Gemini 1.5 Flash} & \includegraphics[height=1.1em]{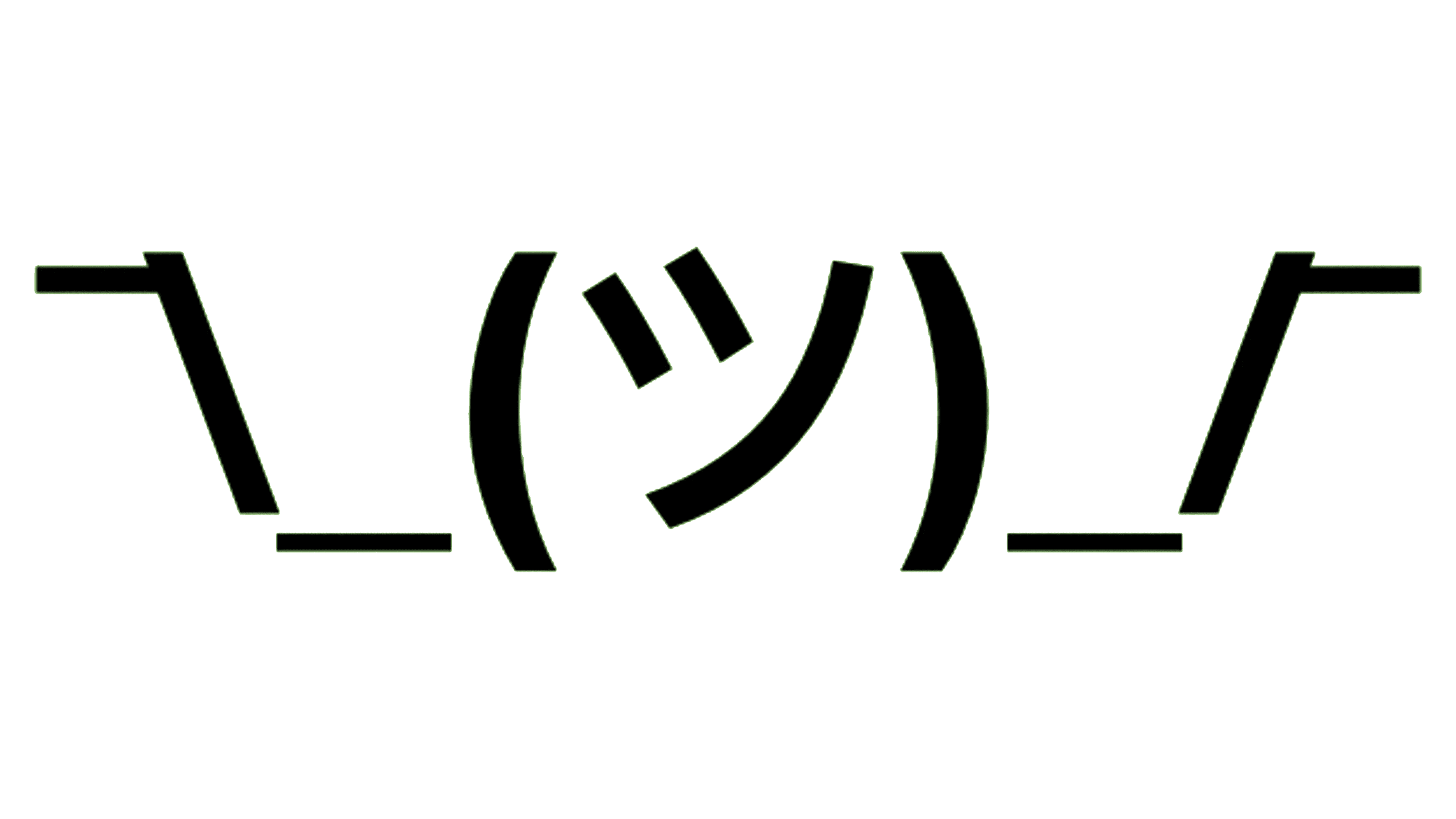}  & 1M & 256K \\
        \faLock\ \href{https://openai.com/index/openai-o3-mini/}{o3-mini-high} & \includegraphics[height=1.1em]{figures/new/shrug_kaomoji.png}  & 200K & 200K \\
        \midrule
       \faUnlock\ \href{https://github.com/deepseek-ai/DeepSeek-R1}{Deepseek-R1}   & 671B/37B & 128K & 129k \\
        \faUnlock\ \href{https://www.llama.com/docs/model-cards-and-prompt-formats/llama3_3/}{Llama 3.3}  & 70B & 128K & 128K \\
        \faUnlock\ \href{https://www.llama.com/llama3_1/} {Llama 3.1}  & 8B & 128K & 128K \\
        \faUnlock\ \href{https://github.com/QwenLM/Qwen2.5}{Qwen 2.5} & 72B & 128K & 152K \\
        \faUnlock\ \href{https://github.com/QwenLM/Qwen2.5}{Qwen 2.5} & 7B & 128K & 152K \\
        \bottomrule
    \end{tabular}
    }
    \caption{Closed-source (\faLock) and open-weights (\faUnlock) models used in the experiments. We report models' parameters count, claimed context length, and vocabulary size.}
    \vspace{-10pt}
    \label{tab:model_comparison}
\end{wrapfigure}
We evaluated all 26 languages on Gemini 1.5 Flash and o3-mini, utilizing Batch API for o3-mini which provided a 50\% cost reduction. For Deepseek-R1, we used the Fireworks API but limited evaluation to English only due to budget constraints. Please refer to \autoref{tab:model_comparison_appendix} and \autoref{tab:model_comparison} for detailed model configurations and associated costs.
For model configurations, we run inference on open-weights models using greedy decoding with temperature set to 0.0, which we also applied to the Gemini-1.5-Flash model. We evaluate Llama and Qwen with vLLM \citep{Kwon2023EfficientMM}, using 1 A100 for smaller models and 4 A100s for larger ones.

\section{Additional Results}
\label{app_sec:additional_results}

\subsection{Retrieval Results}
\label{app_subsec:additional_retrieval_results}
In this section, we detail additional results for the Needle-in-a-Haystack task. 
\paragraph{Average NIAH performance across all languages} In \autoref{fig:model_lineplot_niah}, we see the performance of each model averaged across all NIAH tasks and all languages, with gemini-1.5-flash outperforming all other models, especially at long context-lengths.

\begin{figure}[htbp]
    \centering
    \resizebox{0.9\columnwidth}{!}{\includegraphics{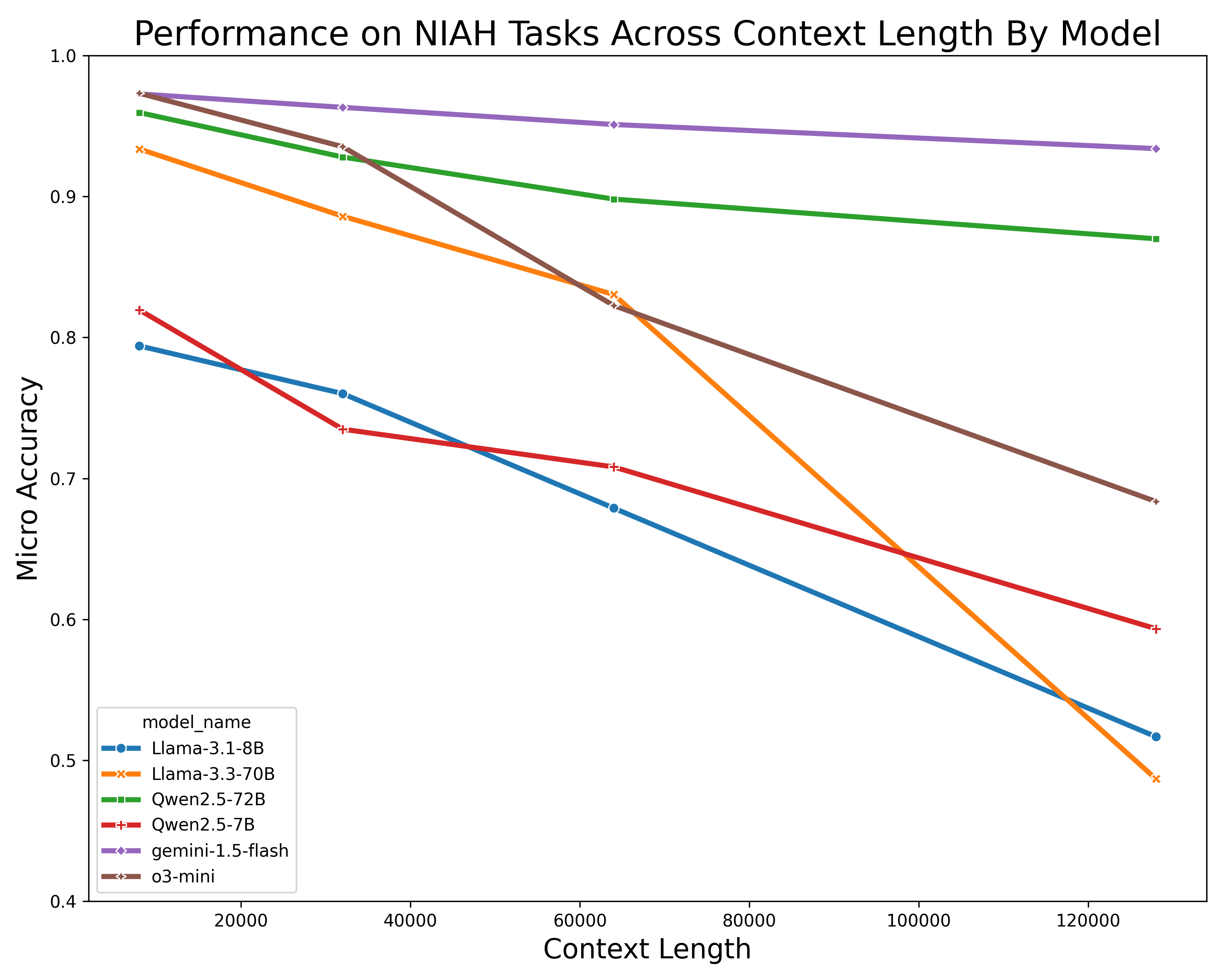}}
    \caption{By model performance across all languages on the NIAH tasks across context lengths. We find that LLama-3.3 has the largest decrease in performance across context lengths.}
    \label{fig:model_lineplot_niah}
\end{figure}

\paragraph{Number of incorrect answers marked as 'none' across all languages} We show that the number of wrong answers labeled as 'none' in S-NIAH tasks with long context varies across models and languages, as illustrated in \autoref{app:fig:wrong_none_closed} and \autoref{app:fig:wrong_none_opensource}. The o3-mini-high model produces a significantly higher number of 'none' errors compared to other models. Interestingly, Gemini-1.5-flash also generates a notable number of 'none' errors for some high-resource languages. Surprisingly, although the Qwen model is specialized in Chinese, it exhibits a large number of 'none' errors in S-NIAH as well.

\begin{figure}[htbp]
    \centering
    \begin{subfigure}{\linewidth}
        \centering
        \includegraphics[width=0.85\linewidth]{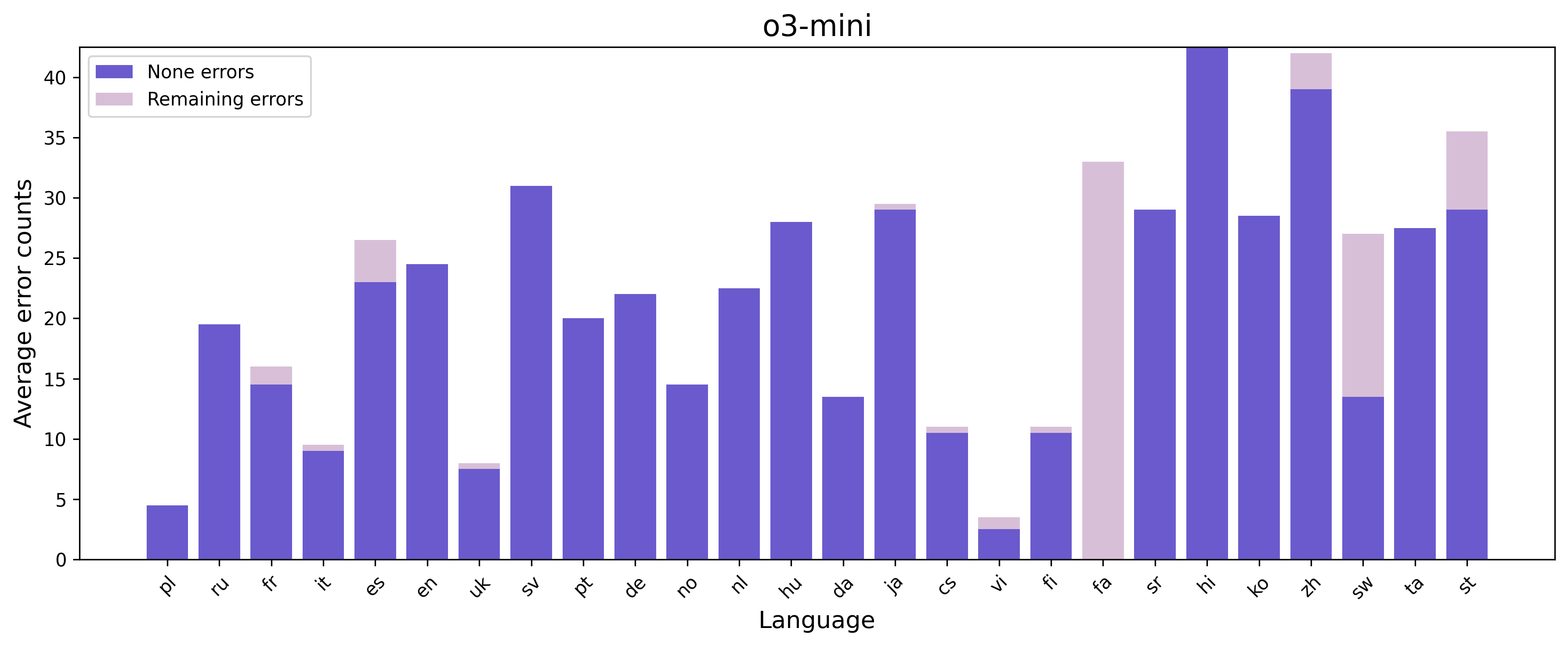}
    \end{subfigure}
    \vspace{1em} 
    \begin{subfigure}{\linewidth}
        \centering
        \includegraphics[width=0.85\linewidth]{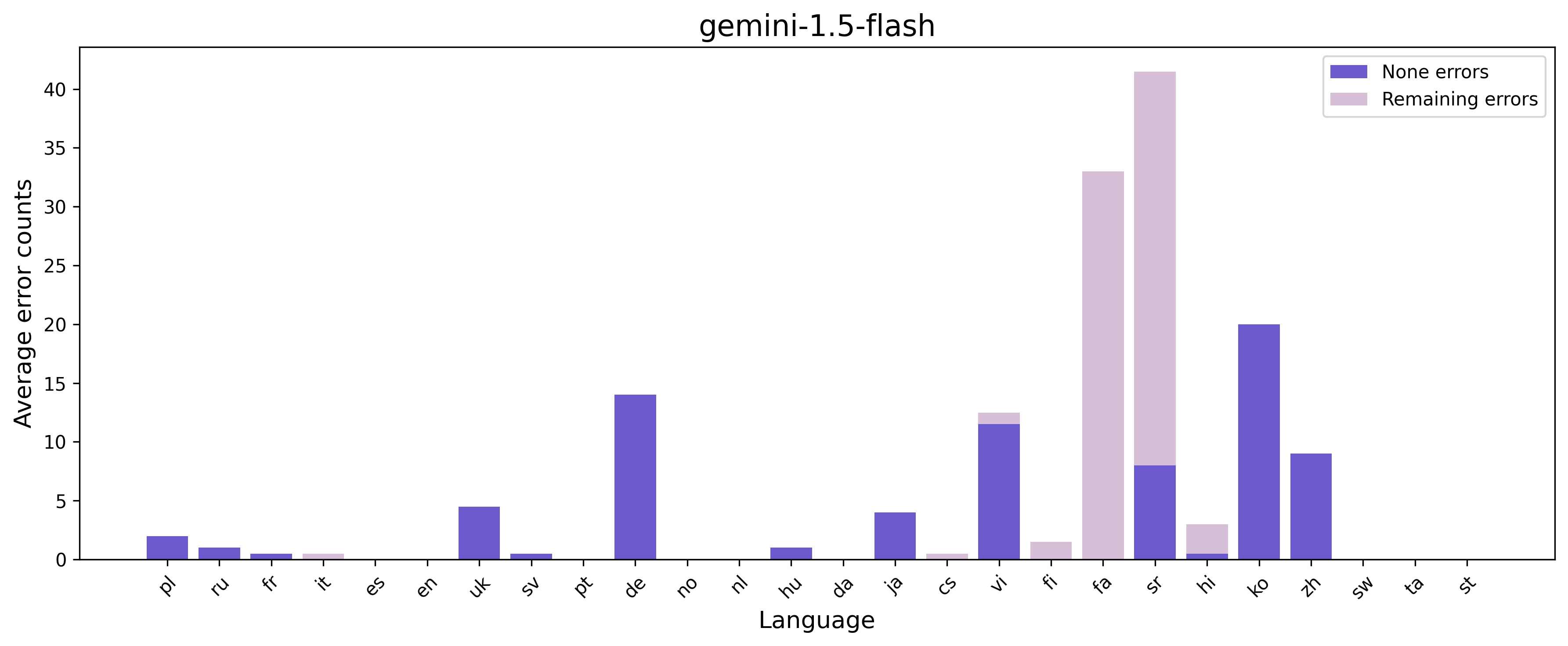}
    \end{subfigure}
    \caption{The types of errors made in the S-NIAH at long-context(64K and 128K) across closed model and languages}
    \label{app:fig:wrong_none_closed}
\end{figure}

\begin{figure}[htbp]
    \centering
    \begin{subfigure}{\linewidth}
        \centering
        \includegraphics[width=0.85\linewidth]{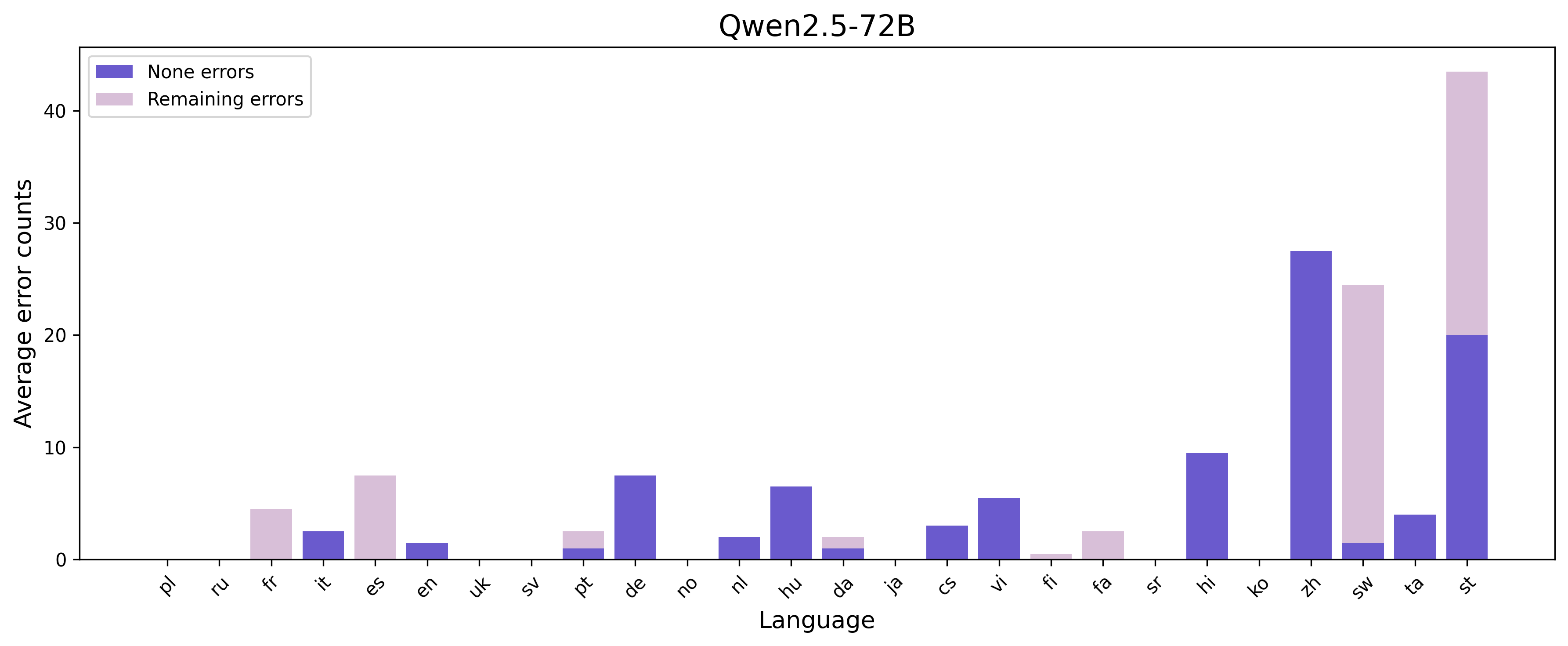}
    \end{subfigure}
    \vspace{1em}
    \begin{subfigure}{\linewidth}
        \centering
        \includegraphics[width=0.85\linewidth]{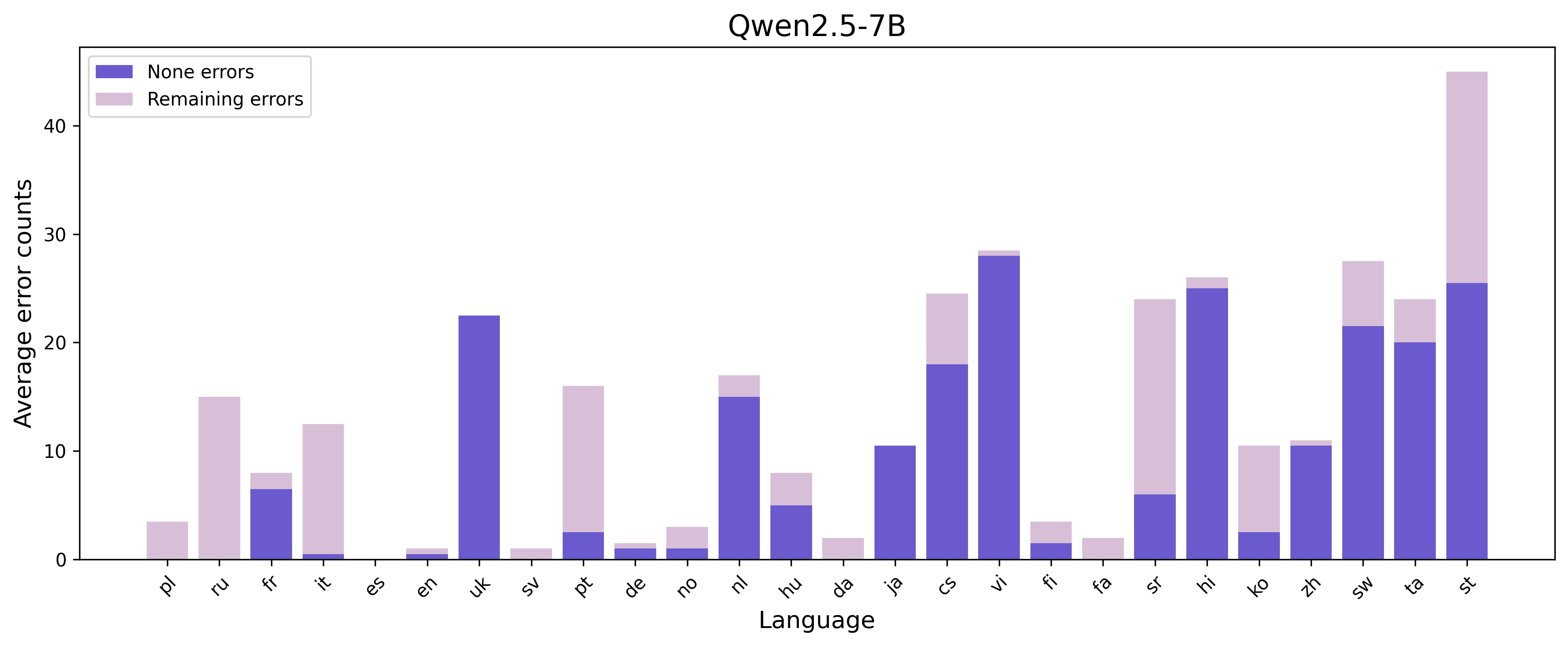}
    \end{subfigure}
    \vspace{1em}
    \begin{subfigure}{\linewidth}
        \centering
        \includegraphics[width=0.85\linewidth]{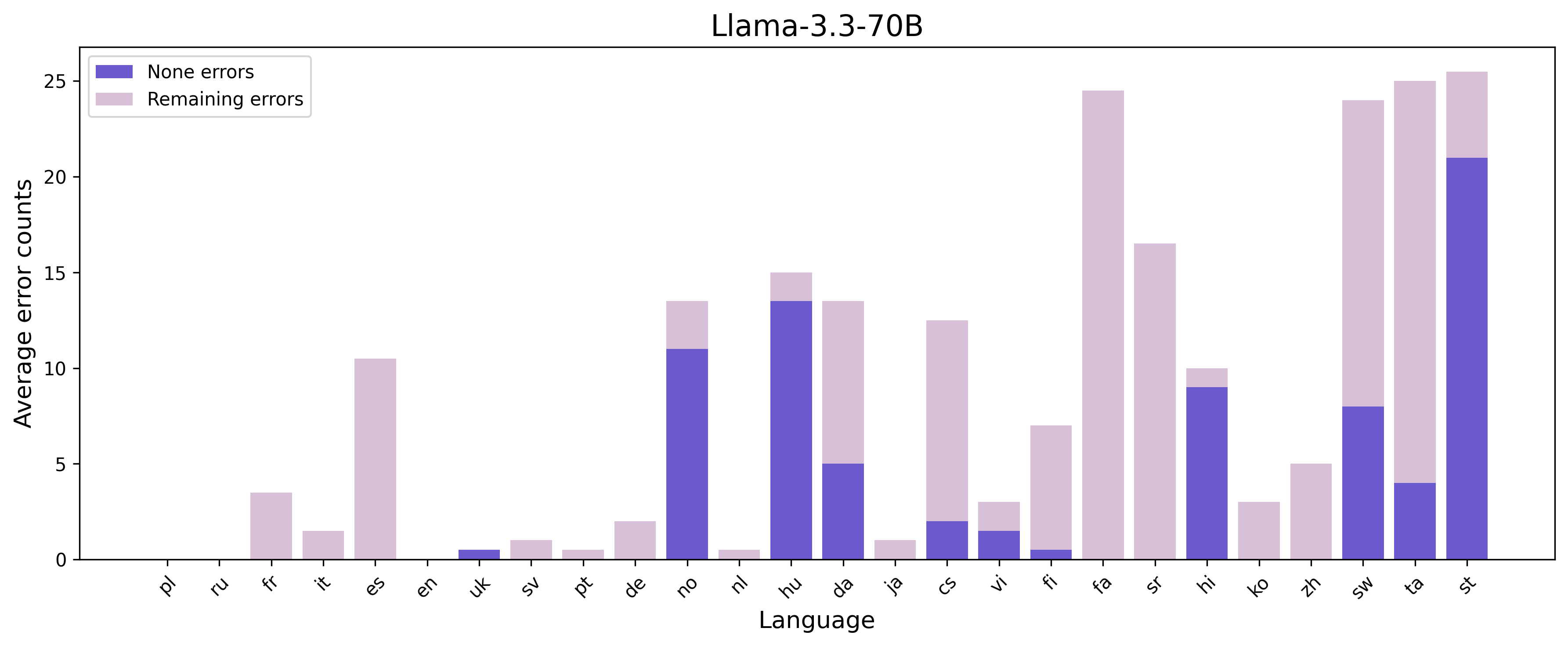}
    \end{subfigure}
    \vspace{1em}
    \begin{subfigure}{\linewidth}
        \centering
        \includegraphics[width=0.85\linewidth]{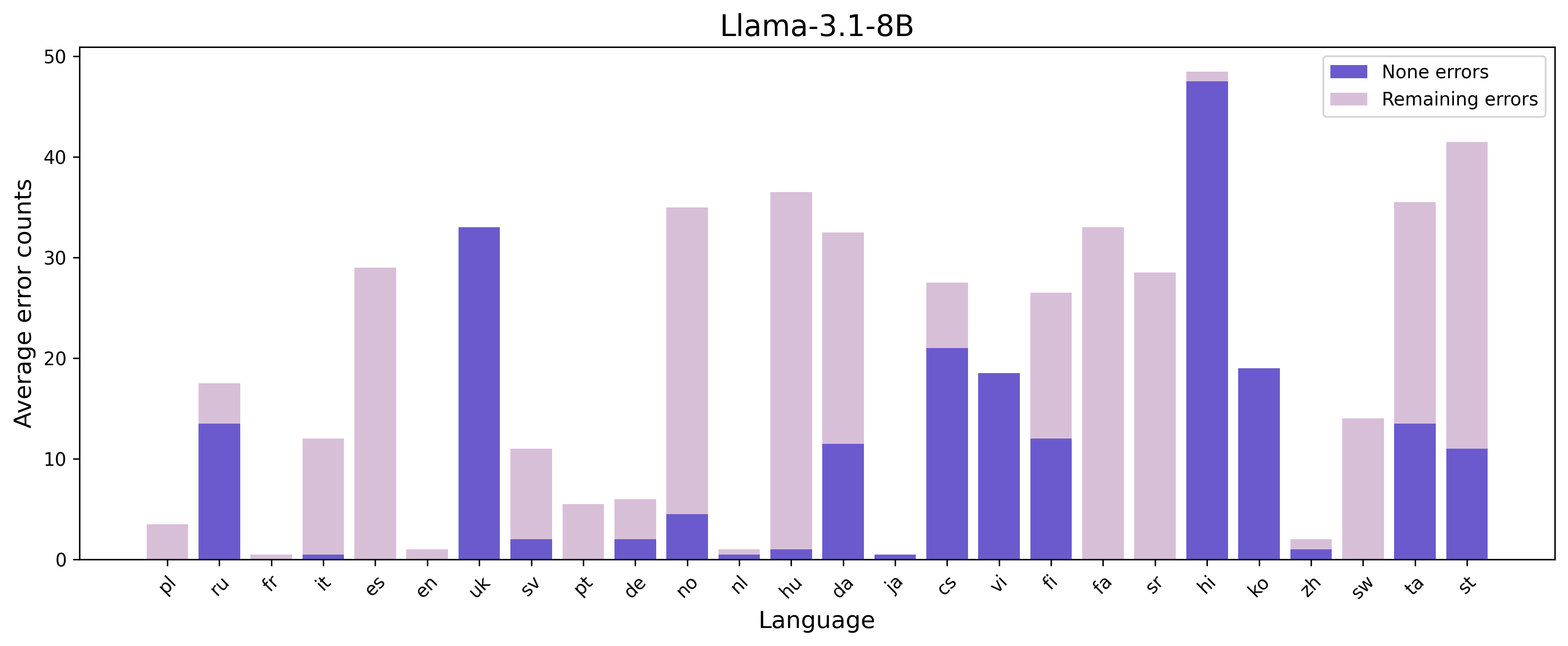}
    \end{subfigure}
    \caption{The types of errors made in the S-NIAH at long-context(64K and 128K) across open-source model and languages}
    \label{app:fig:wrong_none_opensource}
    \end{figure}

\subsection{Reasoning Results}
\label{app_subsec:additional_reasoning_results}
In this section, we detail additional results for reasoning models. In \autoref{fig:avg_reasoning_tokens} we show the average amounts of reasoning tokens for NIAH tasks from o3-mini. We observe that incorrect answers produce drastically higher amounts of reasoning tokens than correct answers. In \autoref{fig:reasoning_error_count}, we show the number of responses exceeding the set max number of output tokens for the English language. We note that both o3-mini-high and deepseek-r1 often exceed the token limit, overreasoning for a simple task.

\begin{figure}[htbp]
    \centering
    \resizebox{0.80\columnwidth}{!}{\includegraphics{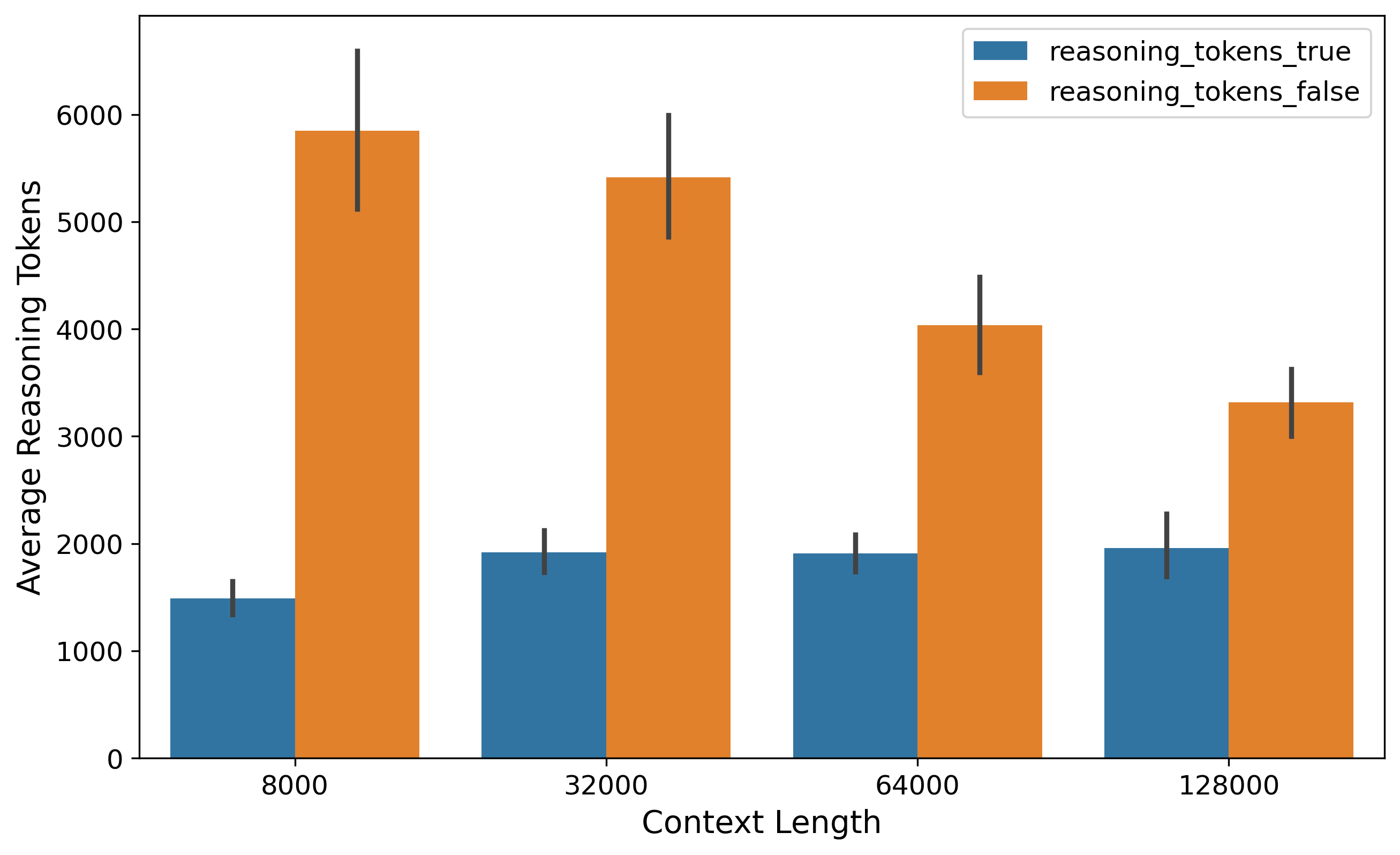}}
    \caption{Average reasoning token length for correct vs. incorrect answers across context lengths in NIAH tasks}
    \label{fig:avg_reasoning_tokens}
\end{figure}

\begin{figure}[t]
    \centering
    \begin{subfigure}[b]{0.49\textwidth}
        \centering
        {\includegraphics[width=\textwidth]{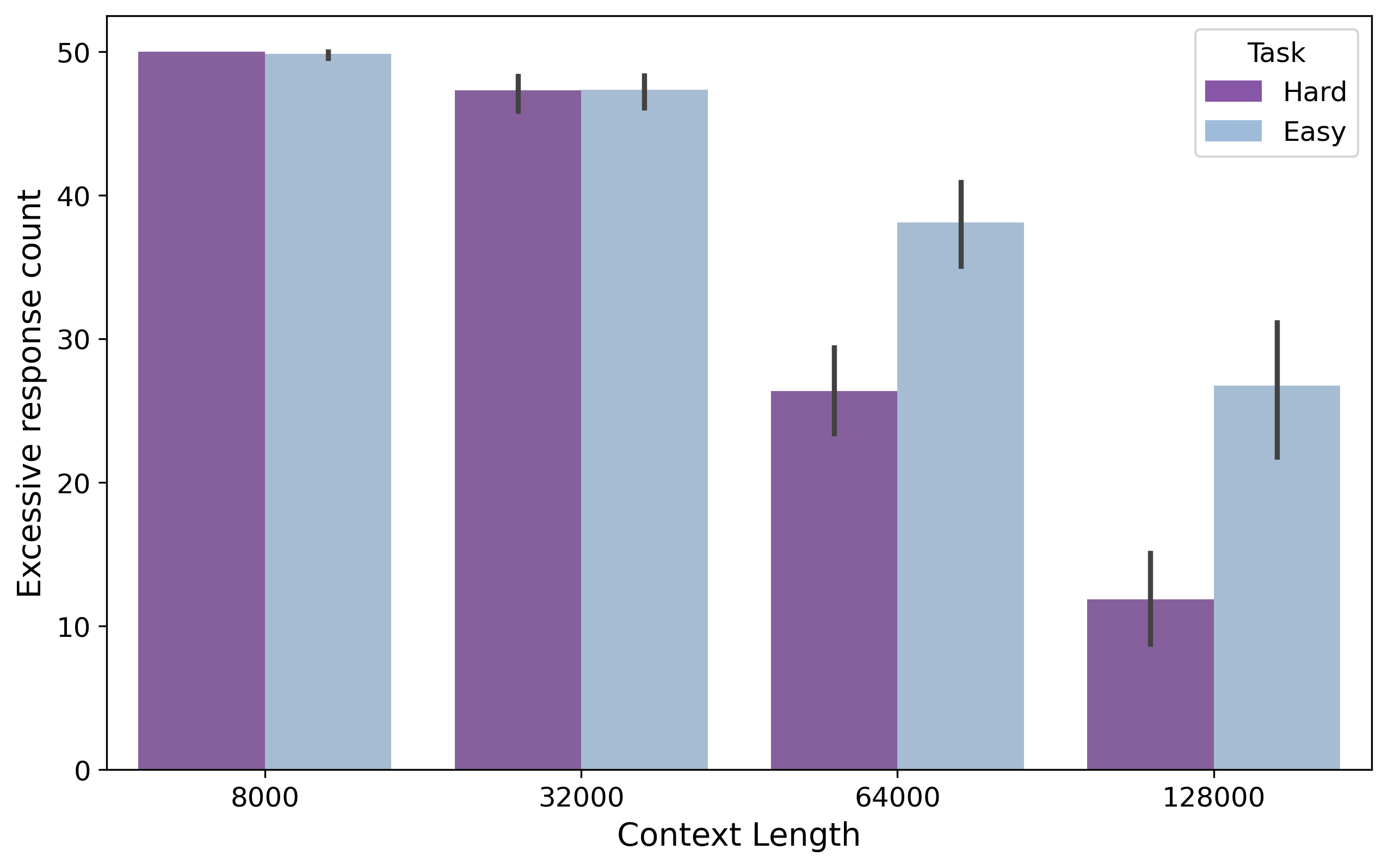}}
        \caption{\textit{o3-mini-high}}
        \label{fig:o3-error}
    \end{subfigure}
    \hfill
    \begin{subfigure}[b]{0.49\textwidth}
        \centering
        {\includegraphics[width=\textwidth]{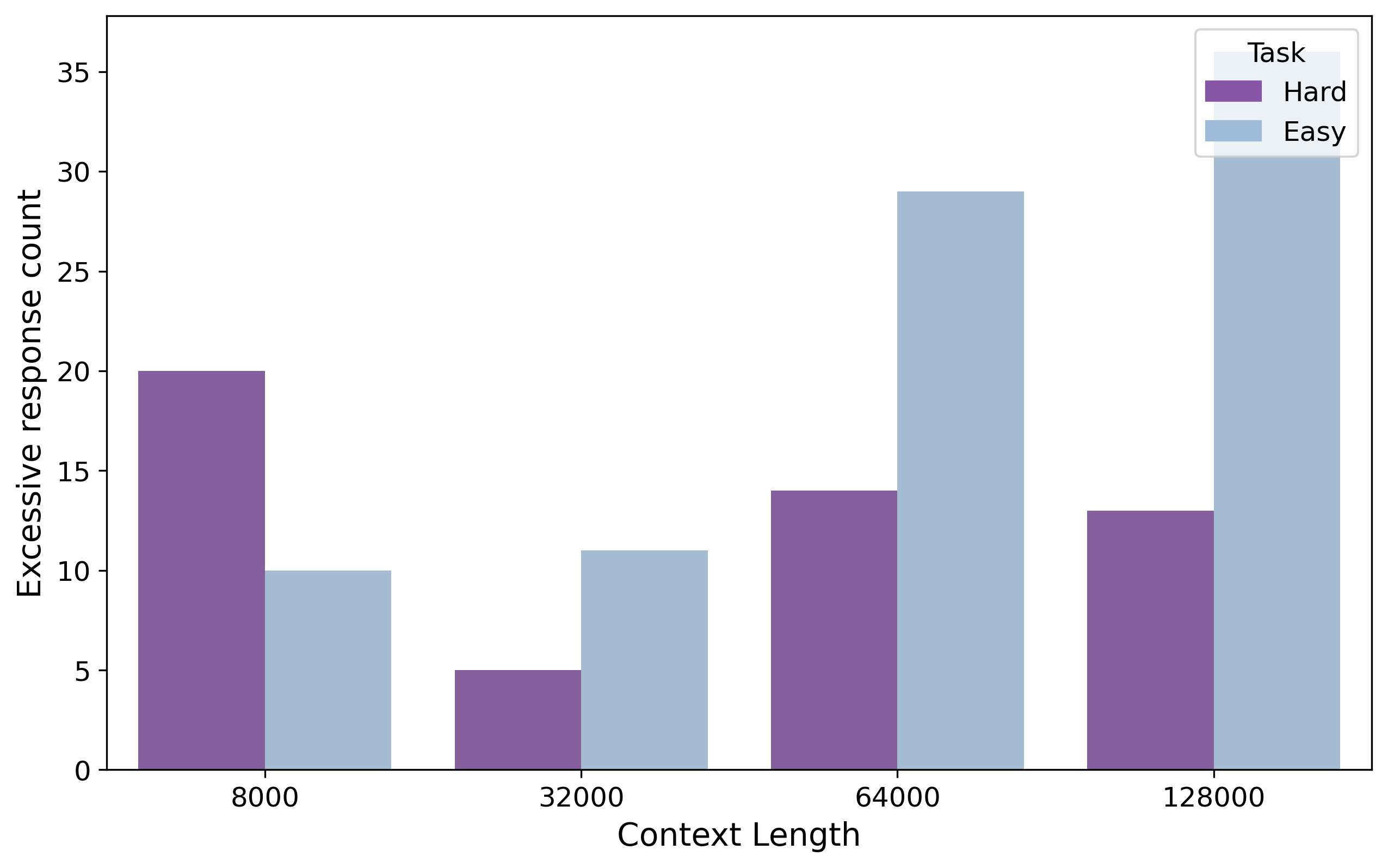}}
        \caption{\textit{Deepseek-R1}}
        \label{fig:ds-error}
    \end{subfigure}
    \caption{Number of responses exceeding the maximum output tokens in \textit{CWE} tasks, caused by reasoning overthinking, for o3-mini-high (26 languages) and Deepseek-R1 (English only).}
    \label{fig:reasoning_error_count}
\end{figure}

\begin{table}[t]
\centering
\begin{tabular}{p{0.95\columnwidth}}
\hline
\textbf{Example of Deepseek-R1's Reasoning for \textsc{MV-Niah}} \\
\hline
\texttt{<think>} \\
Okay, let's tackle this question. The user wants to know the special magic numbers associated with the word ``field'' mentioned in the provided text. I need to go through the text carefully and look for any instances where the word ``field'' is followed by a number.

First, I'll start by recalling that the text is from ``Little Women'' by Louisa May Alcott. The user provided a lengthy excerpt covering several chapters. The task is to find any mention of a ``special magic number for 'field''' in the text. 

I'll need to scan through the text for the exact phrase ``The special magic number for 'field' is: [number].'' Since the user provided the entire text, I can search for occurrences of this specific phrase.

Looking through the chapters, I remember that in some parts, there are notes or numbers inserted. Let me check each chapter section:

1. \textbf{Chapter One}: No mention of a magic number here. The story starts with the sisters talking about Christmas.
2. \textbf{Chapter Two}: Focuses on Christmas morning and the girls' activities. No numbers related to ``field.''
3. \textbf{Chapter Three}: The Laurence boy. Here, there's a mention of a magic number. The text says: ``The special magic number for 'field' is: 2140194.'' So that's one number.
4. \textbf{Chapter Four}: Burdens. No mention here.
5. \textbf{Chapter Five}: Being Neighborly. No numbers here either.
6. \textbf{Chapter Six}: Beth Finds the Palace Beautiful. No magic numbers.
7. \textbf{Chapter Seven}: Amy's Valley of Humiliation. No numbers here.
8. \textbf{Chapter Eight}: Jo Meets Apollyon. No mention.
9. \textbf{Chapter Nine}: Meg Goes to Vanity Fair. Here, another instance: ``The special magic number for 'field' is: 4946066.''
10. \textbf{Chapter Ten}: The P.C. and P.O. No numbers here.
11. \textbf{Chapter Eleven}: Experiments. No mention.
12. \textbf{Chapter Twelve}: Camp Laurence. Another instance: ``The special magic number for 'field' is: 4117625.''

So, in total, there are three instances where the text explicitly states a special magic number for ``field'': 2140194, 4946066, and 4117625. Each of these is found in different chapters. I need to list all three as the answer.
\texttt{</think>} \\
\hline
\end{tabular}
\caption{Example of Deepseek's reasoning for the MV-NIAH task, showing how it systematically analyzes the text by dividing it into chapters and searching for the target pattern.}
\label{tab:deepseek_reasoning_example}
\end{table}

\subsection{Aggregation Results}
\label{app_subsec:additional_aggregation_results}

We evaluate CWE in both the easy and hard settings over four context lengths: 8k, 32k, 64k, and 128k. As seen in \autoref{fig:task-resource}, average English accuracy over all models is only 31.5\% for the CWE-easy task.
The CWE-hard setting proves unsolvable with nearly 0\% accuracy across all models. Performance of each model by context length can be found in \autoref{fig:cwe_acc_by_context_len}. All models evaluated perform much better on the \textit{CWE-easy} task in the 8k length. The drop in performance amidst increasing context length indicates language models still have trouble with this kind of aggregation task in very long-context settings. 
The \textit{CWE-hard} task is near impossible for all models. For a task that would be so simple for humans, the near-total incapability of modern LLMs to complete the task in the harder setting highlights that synthetic tasks like CWE and NIAH, that can be made more challenging as model performance improves, are necessary for ongoing evaluation of models.

\begin{figure}[htbp]
    \centering
    \begin{subfigure}[b]{0.49\textwidth}
        \centering
        {\includegraphics[width=\textwidth]{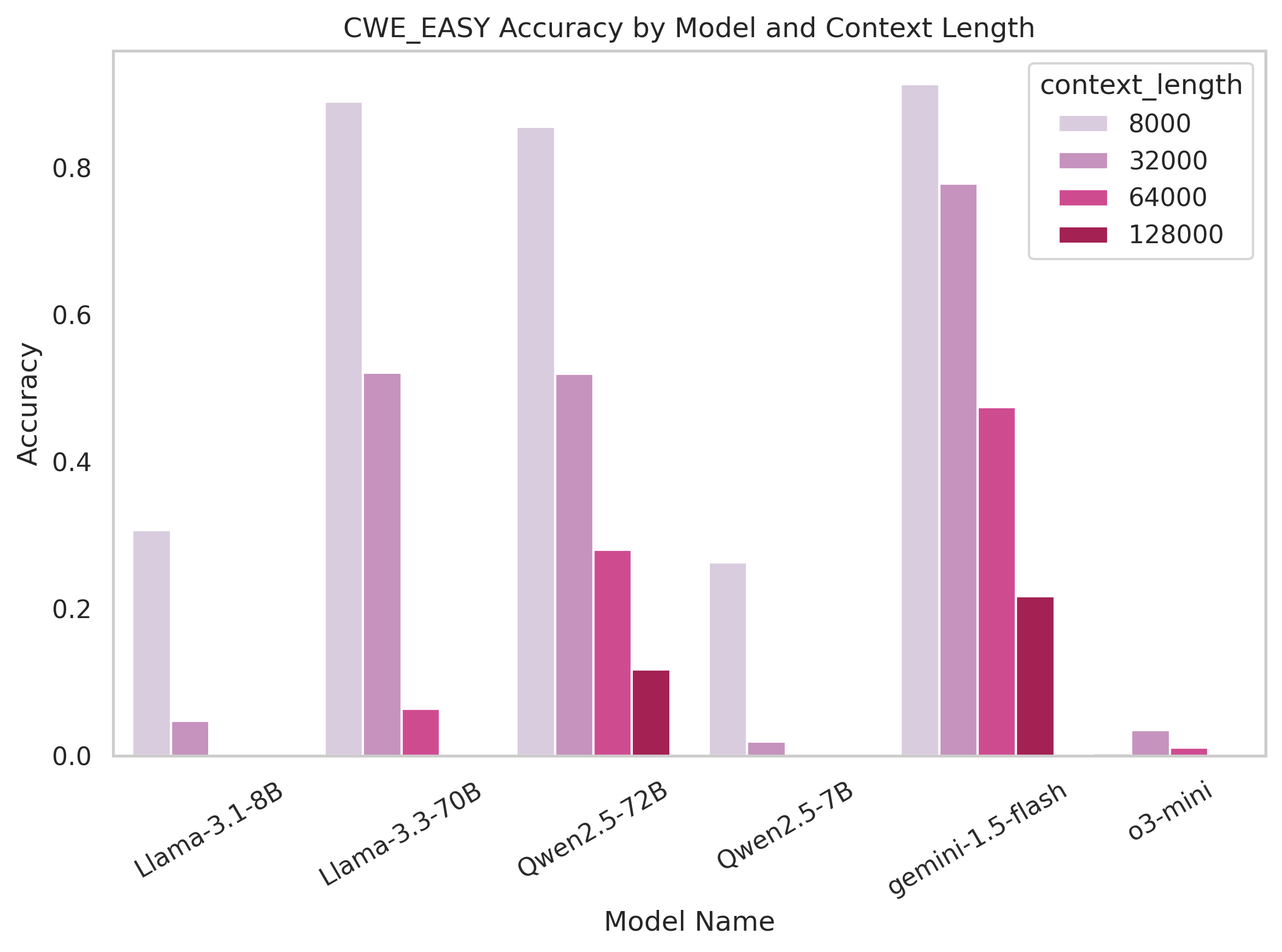}}
        \caption{\textit{CWEeasy}}
        \label{fig:cwe_acc_easy}
    \end{subfigure}
    \hfill
    \begin{subfigure}[b]{0.49\textwidth}
        \centering
        {\includegraphics[width=\textwidth]{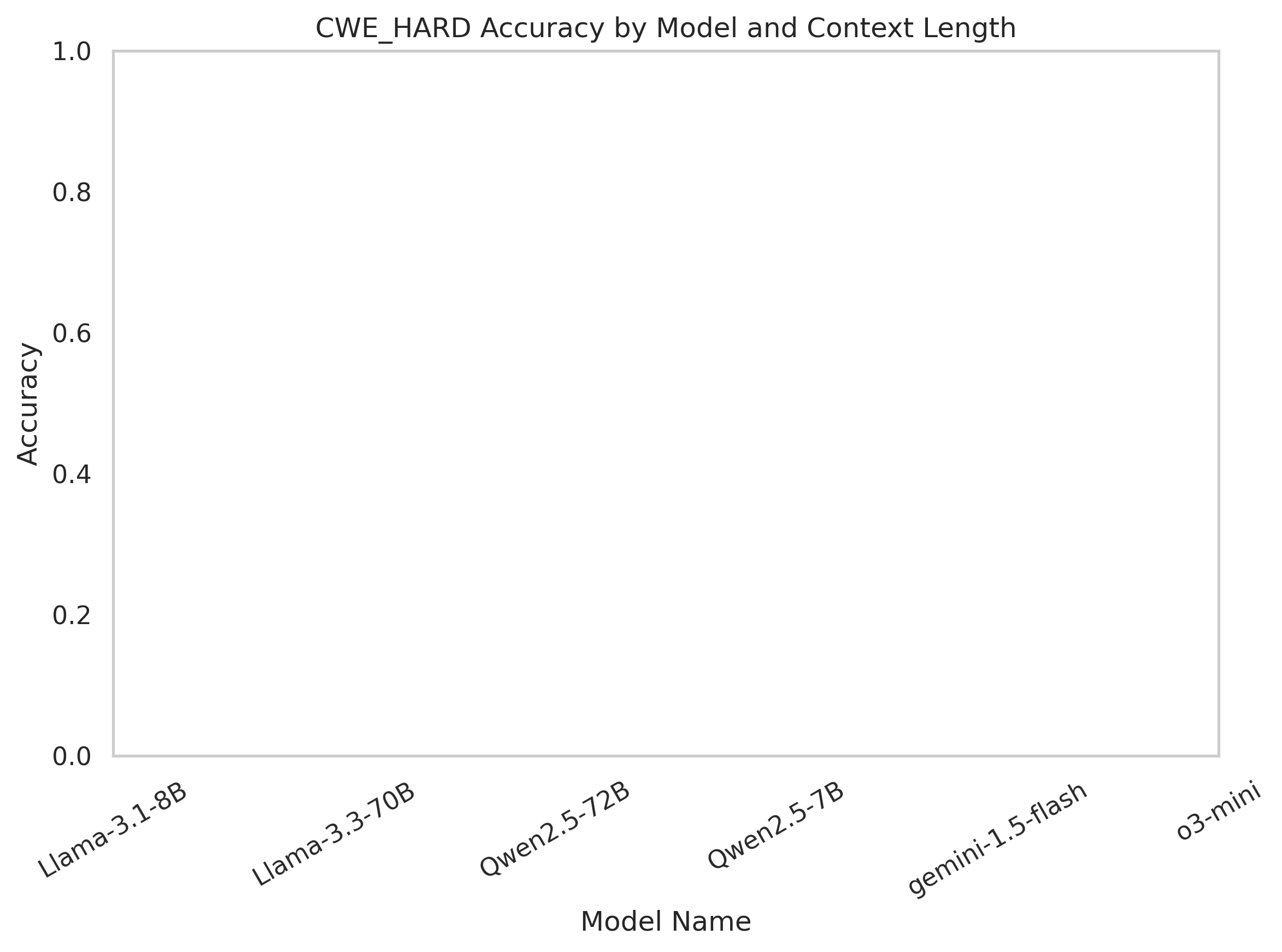}}
        \caption{\textit{CWE-hard}}
        \label{fig:cwe_acc_hard}
    \end{subfigure}
    \caption{CWE Accuracy by Context Length}
    \label{fig:cwe_acc_by_context_len}
\end{figure}


\begin{figure}[t]
    \centering
    \begin{subfigure}[b]{0.49\textwidth}
        \centering
        {\includegraphics[width=\textwidth]{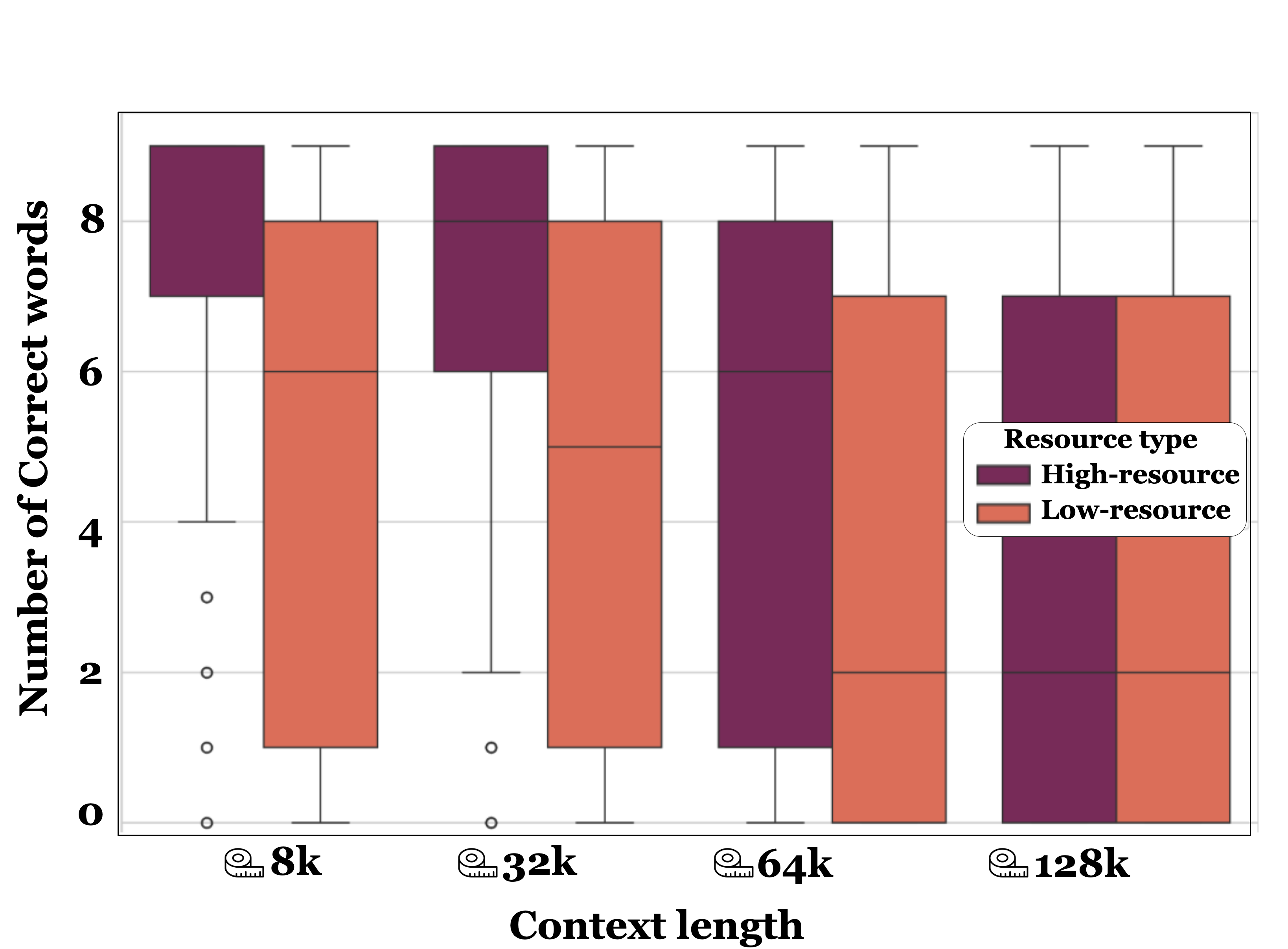}}
        \caption{CWE\_easy task }
        \label{fig:cwe_error_analysis_easy}
    \end{subfigure}
    \hfill
    \begin{subfigure}[b]{0.49\textwidth}
        \centering
        {\includegraphics[width=\textwidth]{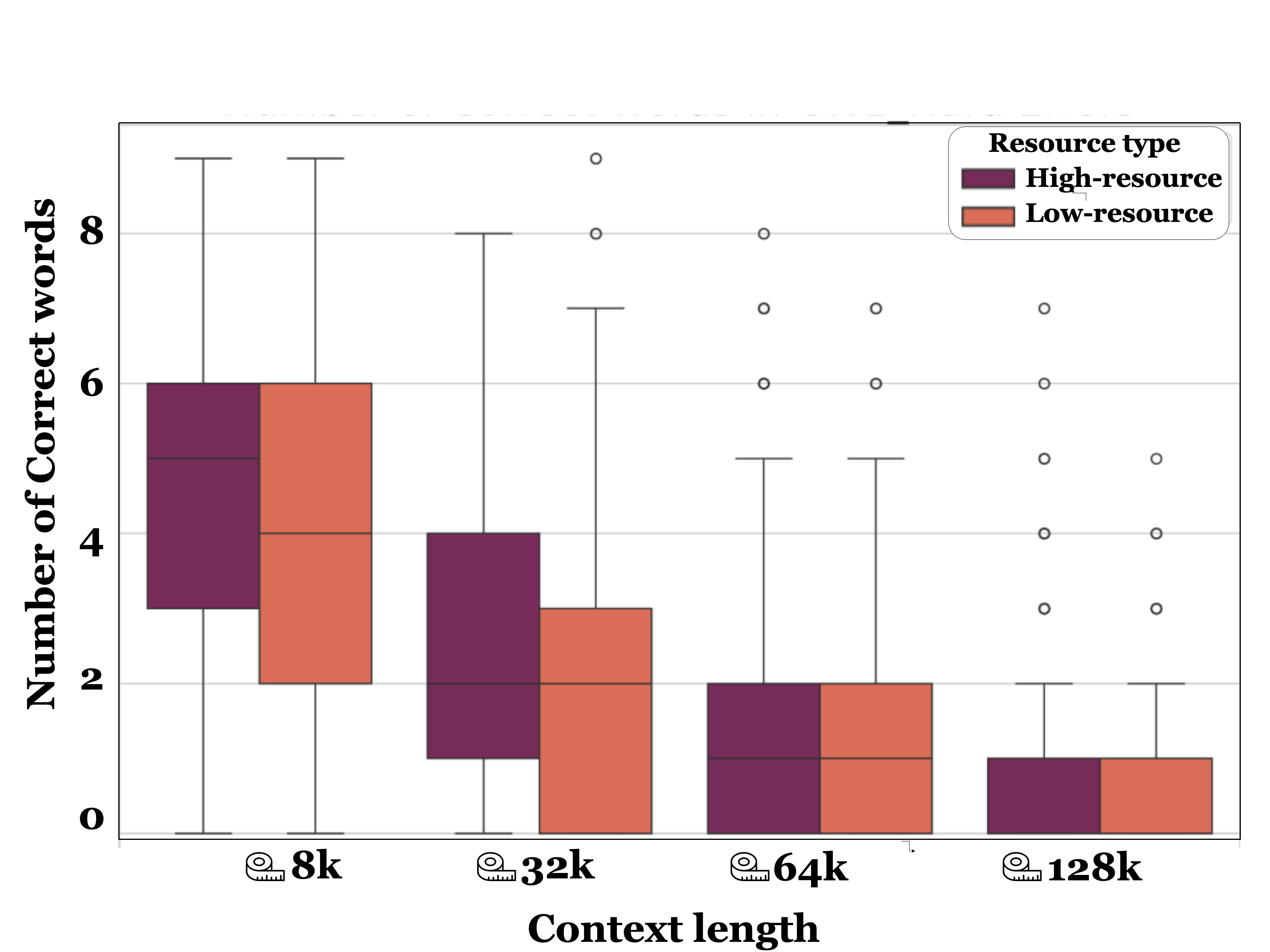}}
        \caption{CWE-hard task}
        \label{fig:cwe_error_analysis_hard}
    \end{subfigure}
    \caption{Number of correct words in cases where the model generated an answer but it was ultimately incorrect at different context length.
    As context length increases, the model finds less correct words. High resource languages tend to return higher average numbers of correct words, which may be indicative of issues relating to tokenization. 
    }
    \label{fig:cwe_error_analysis}
\end{figure}

\paragraph{Reasoning models ``overthink'' on simple aggregation tasks:}
On both CWE tasks, o3-mini-high fails to generate answers within its 10K output token limit
for almost every sample across all languages and context sizes.\footnote{All model configurations reported in \autoref{app_subsec:model_configurations}.}  Notably, this overthinking persists even with smaller contexts, with reasoning outputs sometimes exceeding the length of the given context itself! This is not just unique to o3-mini-high: we also observe that a significant number of Deepseek-R1's responses in the \textit{CWE-hard} task at long contexts exceed the 8k output token limit before reaching an answer as seen in \autoref{fig:reasoning_error_count}. Qualitative inspection of the reasoning tokens reveals strategies unrelated to the task (e.g.,  behaviors (e.g., unrelated to the task. As a whole, these results suggest that reasoning models should be better optimized for tasks requiring aggregating information across long contexts. 

\section{One Tokenizer Does Not Rule Them All}
\label{app_sec:old_experiments}

One of the biggest challenges of the multilingual setting is understanding the effects of the tokenizer on model performance. For example, one model may be able to take in 50,000 tokens in Tamil, but another may only be able to take in 25,000, changing the overall amount of information passed to the model. What would happen if we measured the effects of a model using the information density (or amount of content given), instead of the token count? Below, we detail many of the same experiments as conducted earlier, but in the context of each model being given the same amount of information per model instead of the same amount of tokens.

\paragraph{Drastic Differences in Token Lengths:} For each language, the token counts of other tokenizers when stabilizing for 128k by \texttt{tiktoken cl100k\_base} show astounding variance, as seen in \autoref{fig:token_heatmap}. We see that the same 128k as per \texttt{cl100k\_base} results in large differences between other tokenizers. For example, Tamil has only 33,249 tokens when processed using the \texttt{tiktoken o200k\_base}, compared to 128,000 with \texttt{cl100k\_base}. 

\label{app_subsec:tokenization}
\begin{figure}[htbp]
    \centering
    \resizebox{0.9\columnwidth}{!}{\includegraphics{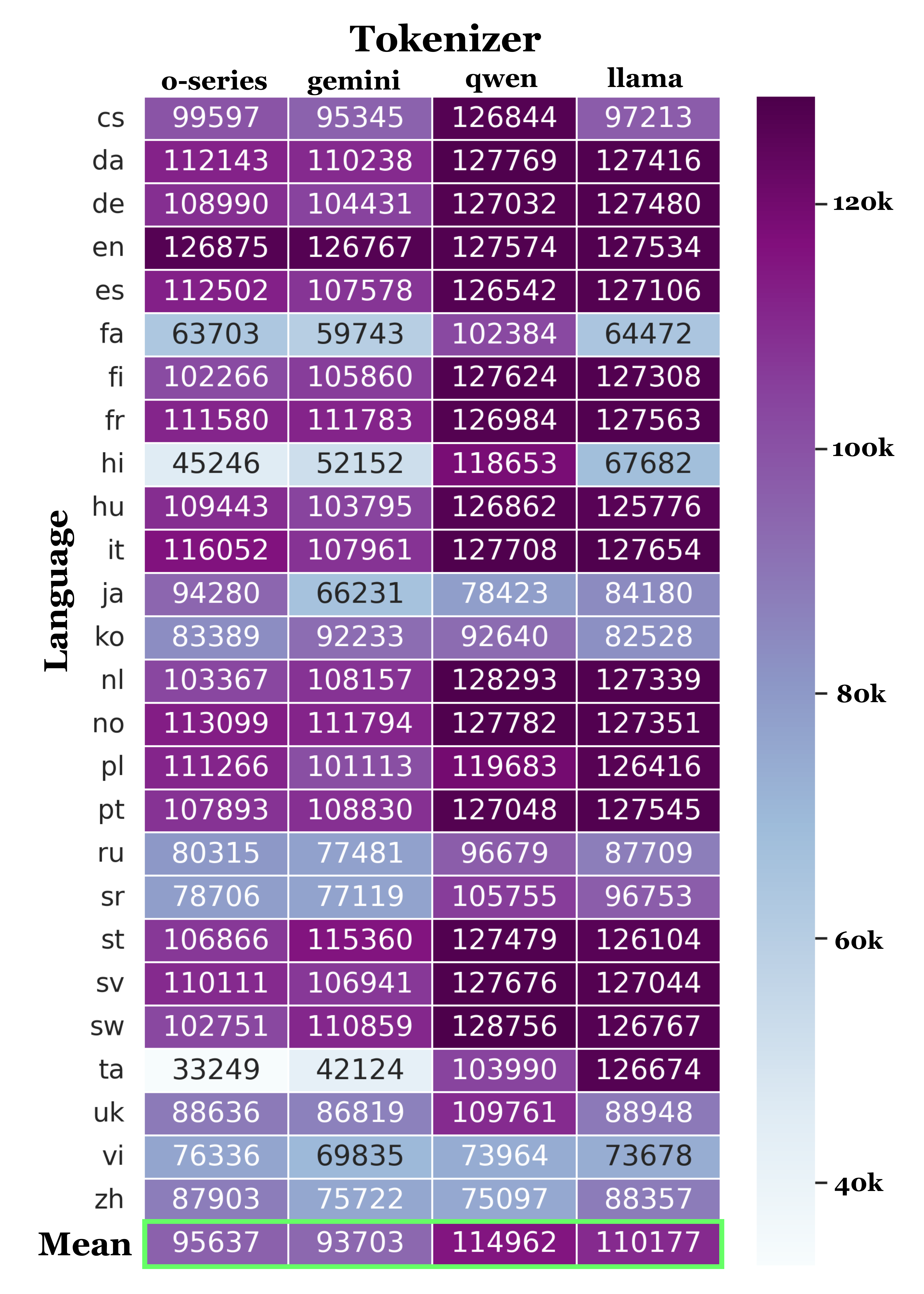}}
    \caption{For each language, the token counts of other tokenizers when stabilizing for 128k by cl100.}
    \label{fig:token_heatmap}
\end{figure}

\paragraph{Similar findings hold despite dramatic differences in tokenized lengths:} As seen in \autoref{fig:old_results_heatmap}, when controlling for amount of context information, Slavic languages (Russian, Ukrainian, Polish) perform highly, surpassing the performance of Romance languages like Italian, French, and Spanish. English performance is even lower, typically 8th. However, similar model ranking patterns remain: Gemini 1.5 Flash and Qwen 2.5 72B perform the best, and high-resource Latin/Cyrillic languages outperform others. It is interesting to note that languages such as Korean, Hindi, and Chinese are towards the low end of the performance spectrum, even though they are often tokenized to shorter lengths when in comparison to English data.\footnote{This is due to larger tokenization gains we have achieved for these languages with newer tokenizer (e.g., \texttt{cl100k\_base} vs \texttt{o200k\_base}.} As context length increases, we still see that the performance gap between high-resource and low-resource languages, consistent with our main results of controlling for number of tokens. 

\begin{figure}[htbp]
    \centering
    \resizebox{0.95\columnwidth}{!}{\includegraphics{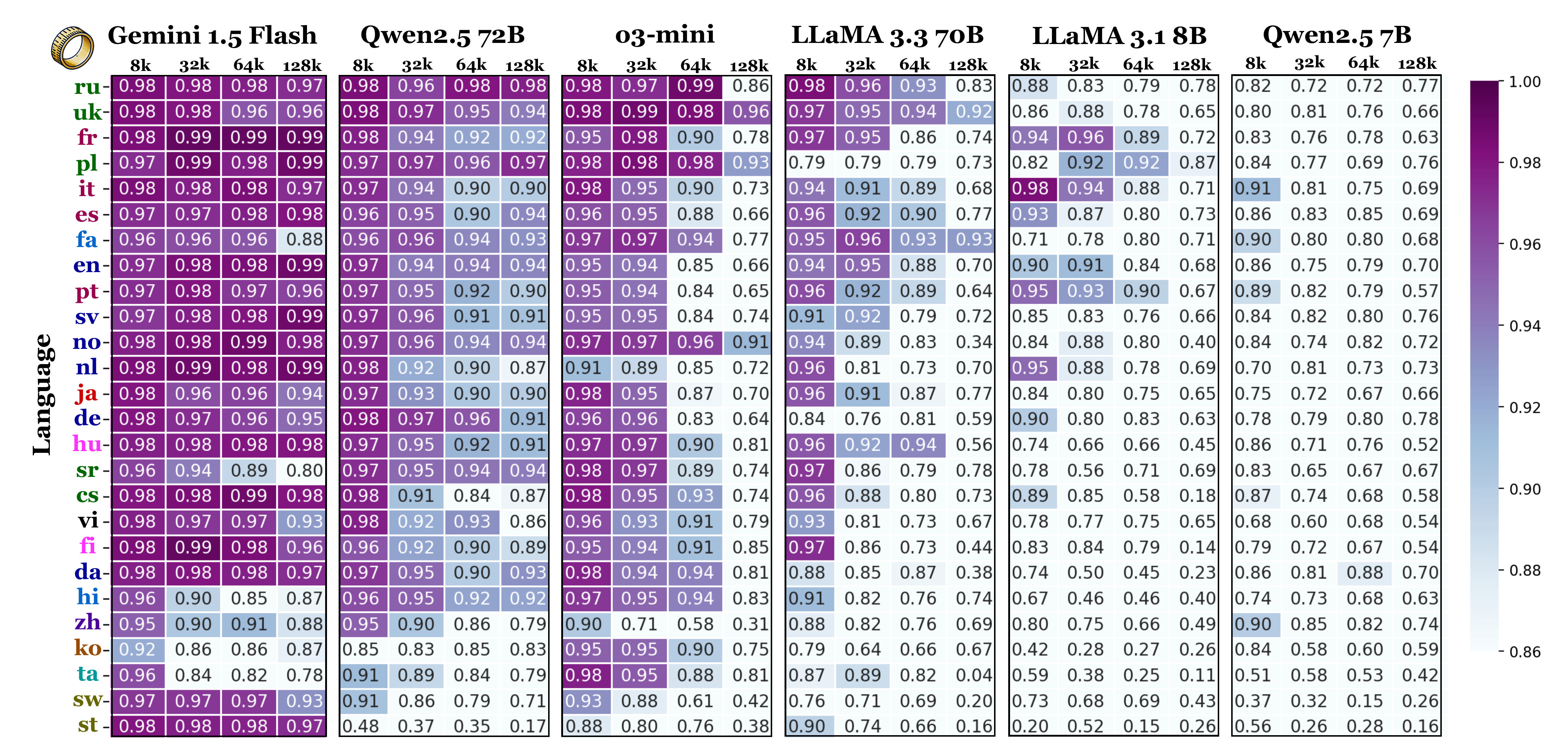}}
    \caption{Heatmap of average accuracy of NIAH tasks by language when controlling for input context length. In these tasks, models are provided with the same input. Input length was measured with \texttt{tiktoken} (cl100k).}
    \label{fig:old_results_heatmap}
\end{figure}

\end{document}